\documentclass[acmtog, nonacm]{acmart}

\usepackage[linesnumbered,ruled,vlined]{algorithm2e}
\usepackage{booktabs} 
\usepackage{pifont}
\usepackage{lipsum}
\usepackage{wasysym}

\usepackage[ruled,vlined]{algorithm2e}
\SetAlCapFnt{\small}
\SetAlCapNameFnt{\small}
\SetAlCapHSkip{0pt}

\acmJournal{TOG}
\newcommand{\revise}[1]{#1}

\usepackage{ulem}
\usepackage{xcolor}
\usepackage{multirow}
\usepackage{makecell}
\usepackage{cleveref}

\crefname{section}{Sec.}{Secs.}
\Crefname{section}{Section}{Sections}
\crefname{table}{Tab.}{Tabs.}
\Crefname{table}{Table}{Tables}
\crefname{figure}{Fig.}{Figs.}
\Crefname{figure}{Figure}{Figures}
\crefname{equation}{Eq.}{Eqs.}
\Crefname{equation}{Equation}{Equations}
\begin{document}

\title{Interspatial Attention for Efficient 4D Human Video Generation}

\author{Ruizhi Shao*$^\ddagger$}
\orcid{0000-0003-2188-1348}
\affiliation{%
 \institution{Tsinghua University}
 \department{Department of Automation}
 \city{Beijing}
 \country{China}
}
\affiliation{%
 \institution{Stanford University}
 \department{Department of Electrical Engineering}
 \city{Stanford}
 \country{United States of America}
}
\email{jia1saurus@gmail.com}

\author{Yinghao Xu*$^\dagger$}
\orcid{0000-0003-2696-9664}
\affiliation{%
 \institution{Stanford University}
 \department{Department of Electrical Engineering}
 \city{Stanford}
 \country{United States of America}
}
\email{justimyhxu@gmail.com}

\author{Yujun Shen}
\orcid{0000-0003-3801-6705}
\affiliation{%
 \institution{Ant Research}
 \city{Hangzhou}
 \country{China}
}
\email{shenyujun0302@gmail.com}

\author{Ceyuan Yang}
\orcid{0000-0003-1417-1938}
\affiliation{%
 \institution{ByteDance Inc.}
 \city{Beijing}
 \country{China}
}
\email{limbo0066@gmail.com}

\author{Yang Zheng}
\orcid{0009-0002-6586-7775}
\affiliation{%
 \institution{Stanford University}
 \department{Department of Electrical Engineering}
 \city{Stanford}
 \country{United States of America}
}
\email{yzheng18@stanford.edu}

\author{Changan Chen}
\orcid{0000-0003-3990-6873}
\affiliation{%
 \institution{Stanford University}
 \department{Department of Electrical Engineering}
 \city{Stanford}
 \country{United States of America}
}
\email{changanvr@gmail.com}

\author{Yebin Liu}
\orcid{0000-0003-3215-0225}
\affiliation{%
 \institution{Tsinghua University}
 \department{Department of Automation}
 \city{Beijing}
 \country{China}
}
\email{liuyebin@mail.tsinghua.edu.cn}

\author{Gordon Wetzstein}
\orcid{0000-0002-9243-6885}
\affiliation{%
 \institution{Stanford University}
 \department{Department of Electrical Engineering}
 \city{Stanford}
 \country{United States of America}
}
\email{gordon.wetzstein@stanford.edu}

\begin{abstract}
Generating photorealistic videos of digital humans in a controllable manner is crucial for a plethora of applications. 
Existing approaches either build on methods that employ template-based 3D representations or emerging video generation models but suffer from poor quality or limited consistency and identity preservation when generating individual or multiple digital humans. 
In this paper, we introduce a new interspatial attention (ISA) mechanism as a scalable building block for modern diffusion transformer (DiT)--based video generation models. ISA is a new type of cross attention that uses relative positional encodings tailored for the generation of human videos. 
Leveraging a custom-developed video variation autoencoder, we train a latent ISA-based diffusion model on a large corpus of video data. 
Our model achieves state-of-the-art performance for 4D human video synthesis, demonstrating remarkable motion consistency and identity preservation while providing precise control of the camera and body poses. 
Our code and model are publicly released at \url{https://dsaurus.github.io/isa4d/}.
\end{abstract}



\begin{teaserfigure}
  \centering  
  \includegraphics[width=0.9\linewidth]{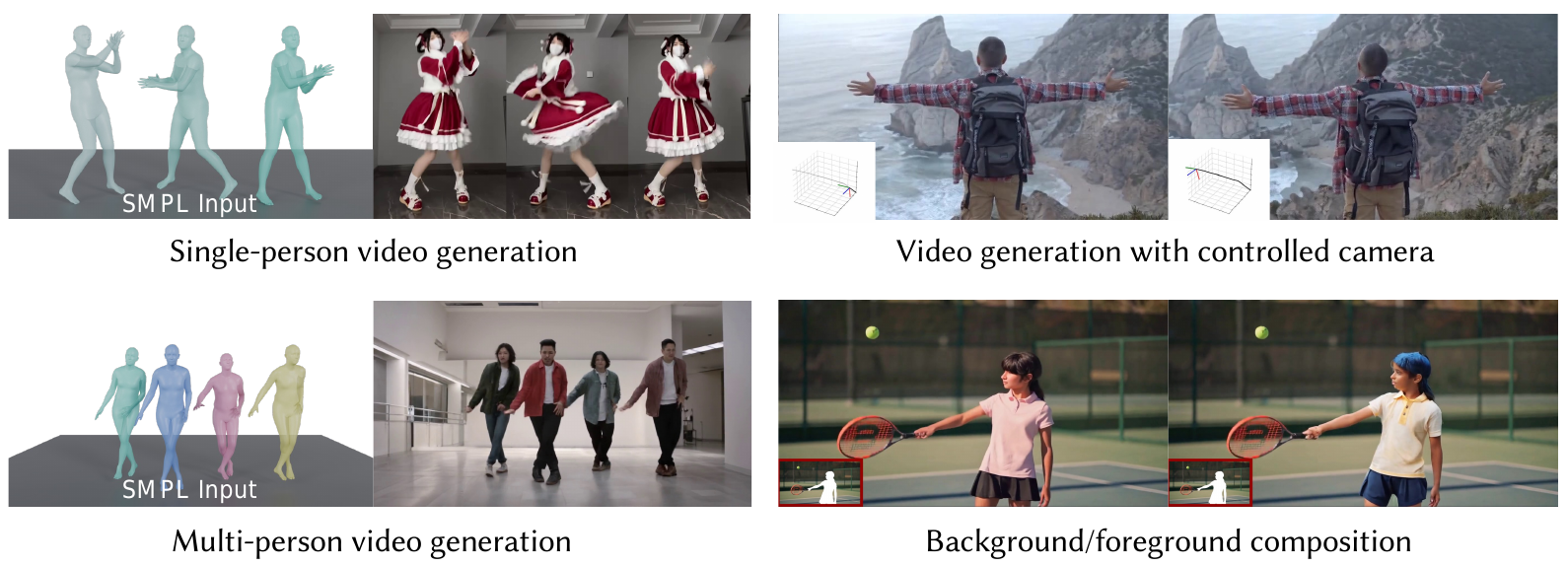}
  \vspace{-5pt}
  \caption{
 We introduce interspatial attention as a building block for diffusion transformer--based generative AI models, enabling high-quality video generation of digital humans with a high level of realism, consistency, and identity preservation. 
 Our 4D-aware model enables a wide spectrum of applications, including single-person and multi-person video generation, video generation with controlled camera trajectory, background/foreground composition, among others.
    }   
\label{fig:teaser}
\end{teaserfigure}

\maketitle

\let\thefootnote\relax\footnotetext{
*Equal contribution $\quad$
$\dagger$ Corresponding author \\ $\quad$
$\ddagger$ Work done during visiting Stanford University
}

\section{Introduction}
\label{sec:intro}

Generating videos of photorealistic humans with full control over camera perspective and body motion is becoming increasingly important for several industries, including visual effects and gaming, teleconferencing, augmented and virtual reality, virtual try-on, robotics, among others. 

To unlock these applications, many existing works model, reconstruct, or generate avatars using parametric template--based representations (see Sec.~\ref{sec:related:template}), including SMPL~\citep{loper2023smpl}. The realism of template-based avatars, however, is often limited as it is challenging for these approaches to model hair and garments, or accurately simulate deformable parts of the avatar. Emerging human video generation models~\citep{hu2023animate,shao2024human4dit,zhu2024champ,xu2023magicanimate}, on the other hand, have shown great promise for controllable generation of photorealistic digital humans. In contrast to template-based approaches, however, video generation methods lack an understanding of the dynamic 3D nature of avatars, as they do not leverage a 3D model or template. This limits multi-frame consistency, identity preservation, the ability to handle multiple characters, and creates other artifacts, for example in the presence of partially occluded body parts. 

We identify two core challenges that limit current human video generation models. First, the variational auto-encoders (VAEs) of recent latent video diffusion models (e.g.,~\citep{yu2023magvit,luo2024open,gupta2023photorealistic,yang2024cogvideox,opensora,zhao2024cvvae}) do not model the fast movements of human motion well, resulting in blurry and low-quality reconstructions and latent representations that hinder the training process of diffusion models of humans.  
Second, current video diffusion models lack explicit 3D parametric human modeling. While previous methods project 3D SMPL models onto 2D planes~\citep{shao2024human4dit, zhu2024champ}, this leads to insufficient geometric cues, making it difficult to handle complex scenarios like self-occlusions and multi-person interactions.

To address the first challenge, we build a video VAE from the ground up. Our VAE introduces spatial and temporal video compression methods, data augmentation strategies, and regularization terms that, together, provide a memory-efficient and high-quality VAE for latent video diffusion models. Our VAE shows noticeably higher-quality reconstructions than alternative approaches for the fast and subtle dynamics of human motion.

Moreover, we introduce a novel and scalable attention mechanism, Interspatial Attention (ISA), that bridges parametric template representations of humans and emerging video diffusion models. Specifically, ISA implicitly builds correspondences between video frames through learnable 3D--2D relative positional encodings in a cross-attention mechanism tailored to digital humans. The key innovation of ISA lies in its symmetric design: the attention module uses tokens extracted from a 3D template used for conditioning the motion as queries, and tokens extracted from 2D video frames as keys/values. This approach effectively propagates 3D features to 2D space for implicit rendering, while the reverse operation propagates 2D features to 3D space, analogous to 3D reconstruction. Our unique bi-directional attention design creates an implicit rendering--reconstruction mechanism within the diffusion transformer. Unlike methods relying on 2D human representations~\cite{hu2023animateanyone,shao2024human4dit,zhu2024champ}, for example via conditioning, ISA enables seamless integration of 3D template representations within the attention module, thereby handling challenging scenarios such as occlusions and multi-person generation. Furthermore, ISA inherits other advantages of the attention mechanism making it seamlessly compatible with state-of-the-art large-scale diffusion transformer architectures.

We design a video diffusion transformer using the proposed ISA blocks and train it on a large corpus of video data for the purpose of 4D human generation with precise control over human motion, camera movement, and background composition. Our method achieves state-of-the-art performance in 4D human video synthesis, generating long videos with consistent motion and appearance across arbitrary viewpoints. Our specific contributions are:
\begin{itemize}
\item We propose a new video VAE design, which facilitates the spatio-temporal compression of videos with fast human motion and builds a well-distributed latent space for diffusion model training.
\item We introduce a novel interspatial attention (ISA) block that facilitates the learning of 3D--2D correspondences for 3D condition injection; ISA can be seamlessly integrated into scalable diffusion transformer architectures for 4D human video generation.
\item We train a video diffusion model using the proposed VAE and ISA mechanisms. Our system achieves state-of-the-art performance in 4D human video synthesis, enabling flexible camera control, multi-character animation, and background composition.
\end{itemize}

\section{Related Work}

\subsection{Template-based Human Animation}
\label{sec:related:template}

Template-based approaches for 4D human animation leverage 3D parametric human representations such as SMPL~\cite{loper2023smpl} in conjunction with efficient neural rendering techniques like NeRF \citep{kolotouros2023dreamhuman}, DMTeT~\cite{huang2023humannorm}, and Gaussian Splatting~\cite{liu2023humangaussian} to generate 3D human animations. These methods, inherently utilizing 3D representations, ensure strict multi-view consistency in the generated animations. Among these, text-to-avatar methods like TADA~\cite{liao2023tada}, HumanGaussian~\cite{liu2023humangaussian}, and HumanNorm~\cite{huang2023humannorm} employ text-to-image diffusion models to optimize a controllable 3D human representation, which is then animated through skinning techniques. However, the dynamic details produced by these methods often lack realism, primarily due to limitations in 3D human body representations and the absence of dynamic priors in text-to-image diffusion models. Alternatively, some methodologies focus on creating personalized avatars through extensive dynamic capture of a specific individual, as exemplified by NeuralActor~\cite{liu2021neuralactor}, HumanNeRF~\cite{weng2022humannerf}, Avatarrex~\cite{zheng2023avatarrex}, and AnimatableGaussian~\cite{li2024animatable}. While these approaches excel in modeling a single person, they suffer from limited generalization capabilities. Moreover, even with the incorporation of generative networks, like GANs~\cite{bergman2022generative,abdal2024gaussian}, these methods are constrained by explicit 3D representations, resulting in dynamic effects that fall short of true photorealism and naturalism. To achieve generalization capability and dynamic realism, our method strategically combines the 3D structural benefits of SMPL with the expressive power of emerging video diffusion models, enabling the generation of realistic and consistent 4D human videos.

\subsection{Video-based Human Animation}
Video models present a promising way for 4D human animation by leveraging deep neural networks, particularly CNNs and Transformers, to directly generate multi-view consistent videos~\cite{sora2024}. These models can implicitly learn spatial relationships and temporal dynamics from video datasets, achieving visually consistent video generations~\cite{valevski2024diffusionmodelsrealtimegame,Dreammachine,RUnway}. Early approaches to human animation primarily focused on 2D image animation based on GANs~\cite{goodfellow2014generative}. GAN-based approaches ~\cite{siarohin2019first, tian2021good, wang2021one, wang2020g3an, siarohin2018deformable, siarohin2019appearance} leverage the generative capabilities of adversarial networks~\cite{goodfellow2014generative, mirza2014conditional} to animate human by transforming reference images according to input motion. These methods typically employ warping functions to generate sequential video frames, aiming to fill in missing regions and enhance visually implausible areas within the generated content. While showing promise in dynamic human generation, GAN-based methods often struggle with generalizable motion transfer, particularly when there are significant variations in human identity and scene dynamics between the reference image and the source video, leading to unrealistic visual artifacts and temporal inconsistencies in the synthesized videos.

Diffusion models, known for their superior generation quality and stable controllability, have been successfully applied to human image animation ~\cite{bhunia2023person, karras2023dreampose, wang2023disco, hu2023animate, xu2023magicanimate, zhu2024champ, shao2024human4dit}. These models employ various strategies, such as reference cross-attention blocks and optical flow in latent space, to enhance the visual fidelity and consistency of generated videos. For instance, Animate Anyone~\cite{hu2023animate}  employs a UNet-based ReferenceNet to inject features from reference images, and incorporates human motion through pose guidance network. While diffusion models achieve reaslitic and high-quality 2D video generation, these methods struggle to generate multi-view consistent videos with physically correct content. The key challenge in extending these models to 4D video generation lies in effectively incorporating 3D conditions. Recent works have begun exploring the injection of camera conditions~\cite{yang2024direct, he2024cameractrl} and 3D SMPL~\cite{zhu2024champ, shao2024human4dit}.  
Champ~\cite{zhu2024champ} employs SMPL as an enhanced animation condition to preserve 3D shape identity and achieve improved human motion control. Human4DiT~\cite{shao2024human4dit}  further leverage a diffusion transformer with temporal and view transformers, simultaneously incorporating 3D SMPL and cameras for enhanced 4D human video generation. However, these methods typically require rendering 3D SMPL into 2D maps, such as normal maps, resulting in the loss of 3D structural information during camera projection. This limitation makes it particularly challenging to handle self-occlusion and multi-person video generation scenarios. To address this challenge, we propose interspatial attention, which efficiently builds the explicit correspondences between 3D SMPL and 2D videos.

\subsection{Variational Autoencoder}
Recent advances in two-stage generative model pipelines have highlighted the crucial role of VAEs~\cite{kingma2013auto} in compressing 2D signals into latent space. Early approaches focused on discrete codebook compression, as pioneered by VQ-VAE~\cite{van2017neural} and enhanced by VQ-GAN~\cite{esser2021taming} and ViT-VQGAN~\cite{yu2021vector} through adversarial training and transformer architectures, but suffered from limited reconstruction quality due to discrete tokens. Later works such as 3D-VQVAE~\cite{yan2021videogpt} and 3D-VQGAN~\cite{ge2022long, yu2023magvit} extended the discrete compression framework to the video domain. Building upon this, MAGVITV2~\cite{yu2023language} and CViViT~\cite{villegas2022phenaki} further introduced causal 3D convolutions and transformers to enable arbitrary-length video compression, yet the discrete token space remained a fundamental limitation for generation quality. In parallel, continuous latent space methods emerged, with CV-VAE~\cite{zhao2024cvvae} and W.A.L.T.~\cite{gupta2023photorealistic} demonstrating impressive results on general video content through 3D VAE architectures and causal temporal modeling. 
Recently, large-scale video models, including CogVideoX~\cite{yang2024cogvideox}, Mochi~\cite{mochi}, and Cosmos~\cite{cosmos}, have extended this approach by developing their VideoVAEs for video compression.
However, these methods struggle with human videos due to their deformable and articulated nature, where fast local and global motions lead to poor reconstruction quality and suboptimal latent distributions. Our work addresses these limitations through advanced data augmentation and latent regularization specifically designed for fast human video compression, facilitating high-quality diffusion model training.

\section{Overview}
\label{sec:overview}

In the remainder of this paper, we first detail the design and training of our video VAE in Sec.~\ref{sec:VAE}. Then, we briefly review basic attention mechanisms in Sec.~\ref{sec:attention:review}, before introducing our new interspatial attention in Sec.~\ref{sec:attention:ISA}. In Sec.~\ref{sec:attention:ISADiT}, we discuss how to incorporate interspatial attention into a modern diffusion transformer architecture for human video generation with control over identity, camera pose, and background. 

Our human video generator takes as input animated SMPL poses for each character as well as a reference image, which can be either a photograph or a generated image. We can optionally specify the camera trajectory and the background. The output is a video that adheres to the motions defined by the input SMPL poses and the identity structure of the reference image. 

\section{Video Autoencoder}
\label{sec:VAE}

Latent diffusion models employ variational autoencoders (VAEs) to compress images or videos into compact latent representations that enable computationally efficient generation~\cite{rombach2022high}. However, we find that existing VAEs struggle to capture the rapid and complex dynamics of human motion. To address the limitation, we present a novel VAE that is built from the ground up to effectively encode such complexity in video data. 

Our compression model is inspired by MAGVITV2~\cite{yu2023language}
and W.A.L.T.~\cite{gupta2023photorealistic}, adopting their unified VAE architecture for joint image--video compression with support for videos of arbitrary length. Formally, let $\mathcal{V}=\{\mathbf{v}_i\}_{i=1}^{1+T}$ denote a video clip consisting of $1+T$ frames where each frame $\mathbf{v}_i \in \mathbb{R}^{H\times W \times 3}$. The encoder $\mathrm{E}(\cdot)$ compresses the video into spatio-temporal latent representations $\mathcal{Z}=\{\mathbf{z}_i\}_{i=1}^{1+t}$, with each latent $\mathbf{z}_i \in \mathbb{R}^{h\times w \times c}$. The corresponding decoder $\mathrm{D}(\cdot)$ reconstructs the video frames from the latent representations. To achieve efficient compression, the encoder downsamples spatially by a factor $f_s = H/h = W/w$ and temporally by a factor $f_t = T/t$. By default, we use $f_s = 8$, $f_t = 4$, and set the latent dimension as $c = 16$. In the following, we present the details of the network architecture (\cref{vae:arch}), training strategy (\cref{vae:training}) and the evaluation protocol (\cref{vae:evaluation}) for the proposed VideoVAE.

\subsection{Architecture}
\label{vae:arch}
\begin{figure}
\includegraphics[width=\linewidth]{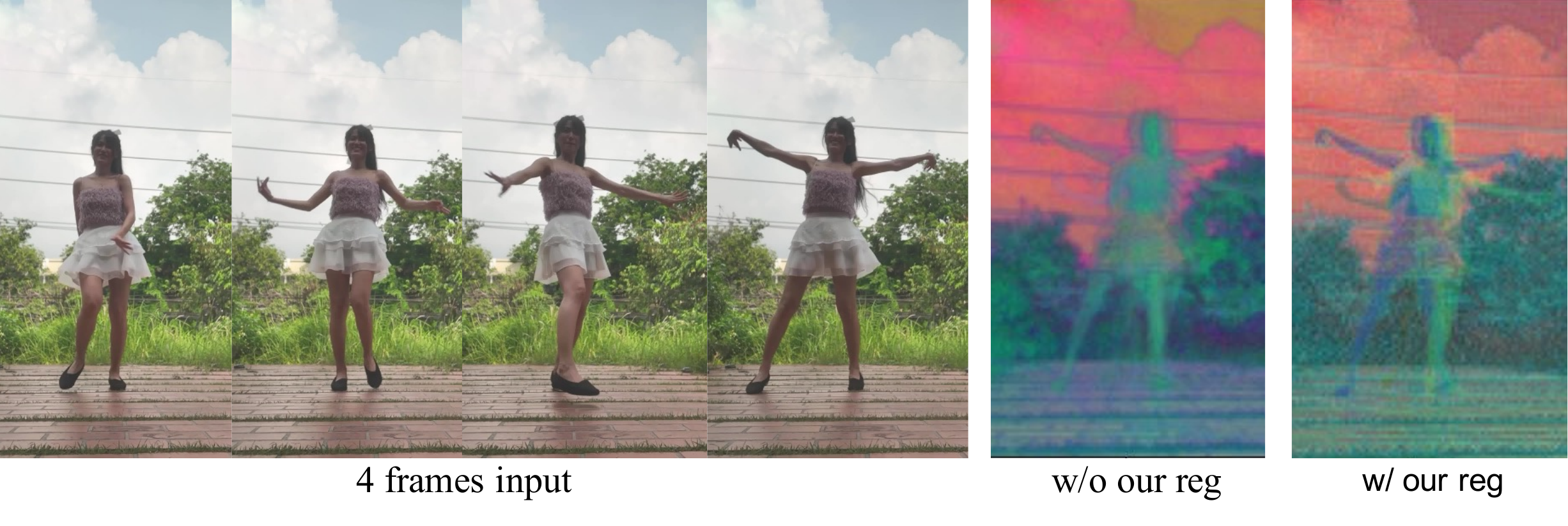}
  \caption{\textbf{Last-frame bias}. The latent tends to compress the final key frame in each temporal window (center right). After adding the \textit{image-decoding regularization}, the latent maintains balanced temporal information distribution across frames (right).}
  \label{fig:last-frame}
\end{figure}

We extend the pretrained image VAE from Stable Diffusion 3 (SD3) \cite{sd3} into a 3D architecture to model temporal dynamics in videos.
The original SD3 VAE architecture consists of cascaded residual blocks interleaved with downsampling (average pooling) and upsampling (resizing plus convolution) layers.
To enable video compression, we inflate this 2D architecture by extending all convolutions to include a temporal dimension, transforming them into 3D convolutions.
For joint image--video compression, we replace regular 3D convolutions with temporally causal 3D convolutions, similar to MAGVITV2~\cite{yu2023language} and W.A.L.T.~\cite{gupta2023photorealistic}.
The causal 3D convolutions ensure that each frame depends only on previous frames, allowing the model to handle both single images and videos of arbitrary length.
Following the SD3 VAE, we enhance reconstruction quality using adversarial losses from a discriminator. While typical image compression frameworks only supervise individual frames, we introduce a 3D discriminator by replacing 2D convolutions with 3D convolutions, thereby capturing temporal dynamics in the reconstructed video.
Despite these modifications, we observe that the trained video VAE model still struggles with fast, articulated human motions, and the spatio-temporal latents show suboptimal distributions as demonstrated in~\cref{fig:last-frame} that hinder the subsequent diffusion training. We thus propose novel training strategies (\cref{vae:training}) to address these limitations.

\subsection{Training}
\label{vae:training}

Next, we introduce two novel training strategies: spatio-temporal data augmentation and image-decoding regularization, designed to achieve  high reconstruction fidelity  and well-structured latent representations for complex human videos.

\paragraph{\bf{Spatio-temporal Data Augmentation.}}
Human motion in videos is inherently challenging to model due to frequent self-occlusions, complex human body and garment deformations, and motion blur.
To tackle these challenges, we introduce two complementary data augmentation strategies as described below:

\textbf{1)} \textit{Random Structured Motion.} To address large spatial displacements (such as squatting and jumping), 
we randomly translate each video frame in different directions and at varying velocities.
This structured motion perturbation encourages the model to learn how to reconstruct significant spatial shifts, enhancing its robustness in handling challenging global motions.

\begin{figure}[t]
\includegraphics[width=\linewidth]{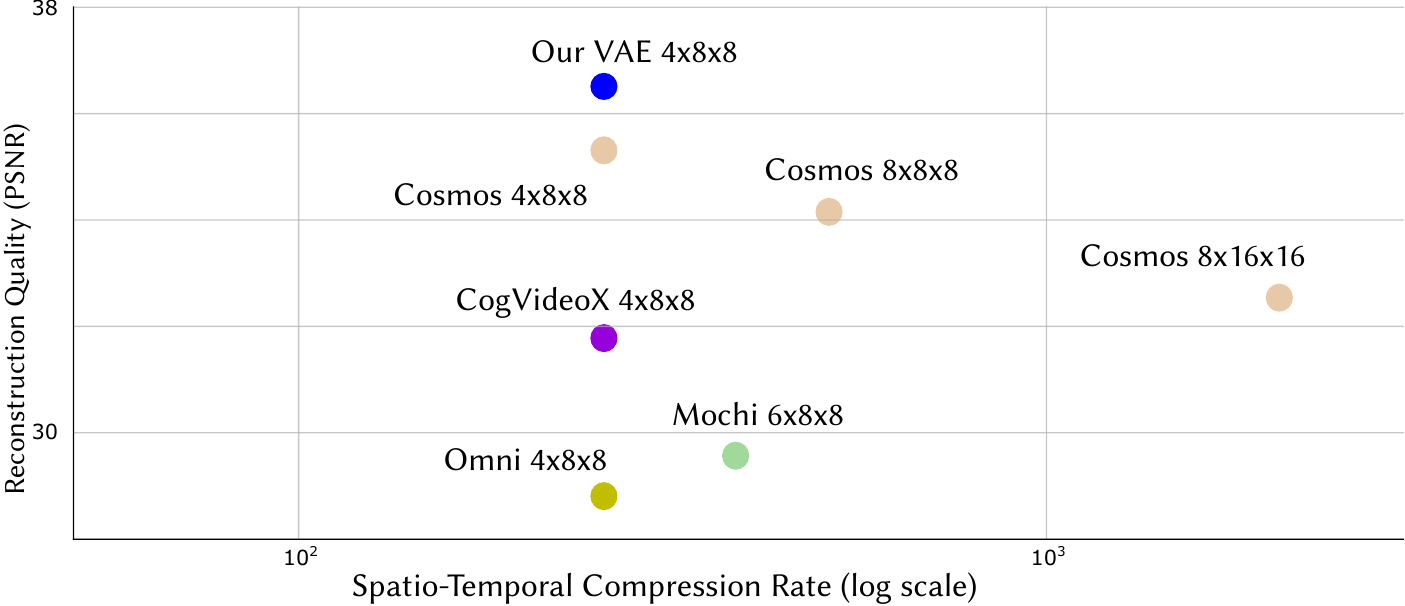}
\caption{\textbf{Comparison of video tokenizers on spatio-temporal compression rate (log scale) vs. reconstruction quality (PSNR).} Each point represents a trade-off between compression ratio and reconstruction quality for different tokenizer configurations. Our method achieves better reconstruction quality than existing video tokenizers. 
}
\label{fig:vae-compare-plot}
\end{figure}

\textbf{2)} \textit{Dynamic Speed Adjustment.} Targeting fast local motions (such as fast hand movements), we modulate video frame rates to generate diverse motion speed samples. This temporal adaptation strategy creates varied temporal densities of motion representation, effectively improving the model's robustness to fast local movements.

\paragraph{\bf{Image-decoding Regularization.}}
The shape of the latent distribution plays a crucial role in the performance of diffusion model training. For example, a dataset whose latent space distribution is irregularly shaped or which has a high variance might be more challenging to be learned by a diffusion model.
While following prior works~\cite{rombach2022high, sd3} to impose a slight KL-penalty to notch the latent distribution towards a normal distribution, we observe a ``last-frame bias'' phenomenon in the learned latents of our video VAE---the latent tends to primarily compress the last frame in each temporal window, as shown in \cref{fig:last-frame}.
This last-frame-biased compression makes the latent distribution suboptimal for diffusion model training, causing severe artifacts during frame transitions at temporal window boundaries in generated videos, especially in videos with fast motion.

\begin{table}[t!]
    \centering
    \caption{\revise{\textbf{Quantitative comparison of video tokenizers}. While omitting regularization terms slightly improves reconstruction quality, adding regularization makes video diffusion model training more efficient. }}
    \begin{tabular}{c|c|c|c|c}
    \toprule
       Method & PSNR$\uparrow$ & SSIM$\uparrow$ & LPIPS$\downarrow$ & FVD$\downarrow$ \\  \hline 
      Mochi & 31.78 & 0.946 & 0.036 & 31.94 \\
Cosmos $4 \! \times \! 8 \! \times \! 8$ & 35.31 & 0.972 & 0.028 & 15.72 \\
CogVideoX & 32.54 & 0.954 & 0.035 & 25.85 \\
Ours (w/o reg) & \textbf{36.71} & 0.980 & \textbf{0.014} & \textbf{11.57} \\
Ours & 36.59 & \textbf{0.981} & 0.015 & 12.16 \\
      \bottomrule
    \end{tabular}
    
    \label{tab:vae_quan_comp}
\end{table}

\begin{figure}[t]
\includegraphics[width=0.75\columnwidth]{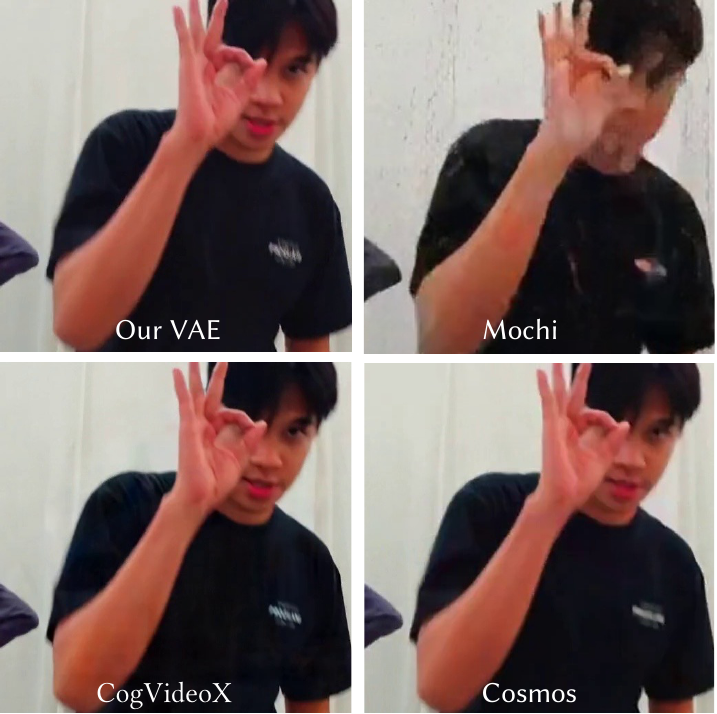}
  \caption{\textbf{Qualitative comparison of video VAEs}. We compare the reconstruction quality of different VAEs using a crop of a 1920$\times$1080 video for our VAE, Mochi, CogVideoX, and Cosmos. Mochi's VAE is noticeably worse than all others with Cosmos also being blurrier than CogVideoX and ours.
  }
  \label{fig:vae-qualitative}
\end{figure}

To address this problem, we introduce an image-decoding regularization term that incorporates an auxiliary image decoder to reconstruct input video frames. Specifically, we decompose each 16-channel latent $\mathbf{z}_i$ into four 4-channel sub-latents, each independently decoding individual frames by the auxiliary image decoder. This frame-wise independent decoding serves as an implicit constraint for balanced temporal information distribution, mitigating the last-frame bias and producing well-structured latents that benefit diffusion training.

\paragraph{\bf{Data and Objectives}}
We utilize the action recognition dataset Kinetics-600~\cite{kay2017kinetics} and the human video generation dataset Human4DiT~\cite{shao2024human4dit}, comprising 600K videos for VAE training.
To enable inference on long videos, we propose a two-stage training scheme: first training our model on short (33 frames) sequences, then fine-tuning it on long (97 frames) sequences. 
We train our video VAE using multiple objectives: the \(L_1\) loss \(\mathcal{L}_{L1}\), perceptual loss \(\mathcal{L}_{p}\), KL divergence loss \(\mathcal{L}_{KL}\), 2D GAN loss \(\mathcal{L}_{2DGAN}\), 3D GAN loss \(\mathcal{L}_{3DGAN}\), and a regularization term for the image decoder \(\mathcal{L}_{reg}\):
\begin{align}
    \mathcal{L} &= \lambda_{L1}\mathcal{L}_{L1} + \lambda_{p} \mathcal{L}_{p} + \lambda_{KL}\mathcal{L}_{KL} + \lambda_{reg}\mathcal{L}_{reg} 
    \\ &+ \lambda_{3DGAN}\mathcal{L}_{3DGAN}  + \lambda_{2DGAN}\mathcal{L}_{2DGAN},  
\end{align}
where $\lambda_{L1}$, $\lambda_{p}$, $\lambda_{KL}$, $\lambda_{reg}$, $\lambda_{2DGAN}$ and $\lambda_{3DGAN}$ are the weights for each respective loss term.

\subsection{Evaluation}
\label{vae:evaluation}

\paragraph{\bf{Data and Metrics.}}
To evaluate the video VAE, we curate an evaluation dataset of 200 high-resolution human videos featuring multi-person interactions, complex textures, and fast motions. We evaluate reconstructed video quality using peak signal-to-noise ratio (PSNR). For comparison, we select state-of-the-art video VAE baselines including the Cosmos tokenizer~\cite{cosmos} and video VAE models from Mochi~\cite{mochi}, and CogVideoX~\cite{yang2024cogvideox}. 

\begin{figure}[t]
\includegraphics[width=0.9\linewidth]{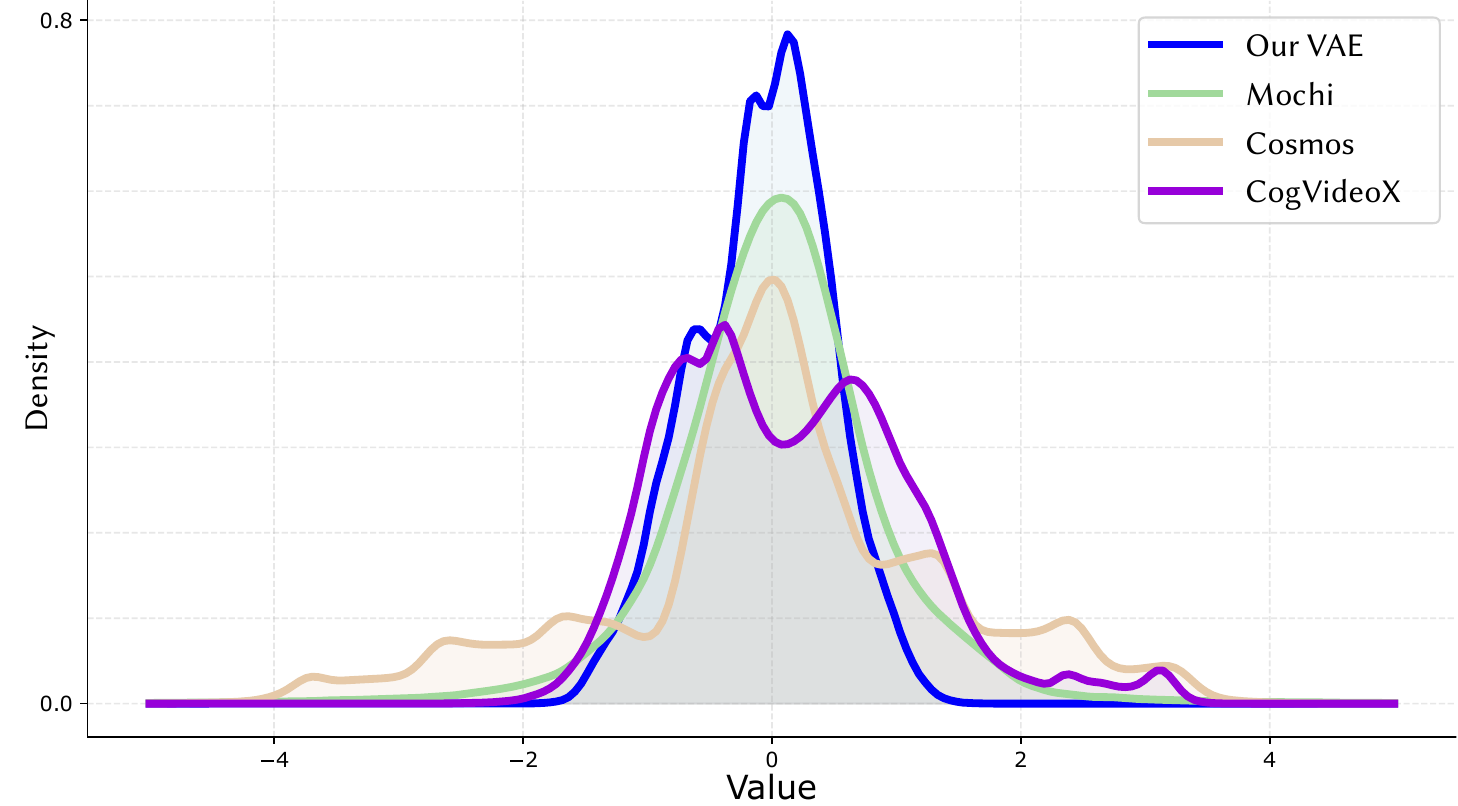}
\caption{\textbf{Comparison of latent distribution from different approaches}. We visualize the latent distributions on the evaluation videos. Our method yields well-structured latent representations compared to baseline methods.}
\label{fig:vae-latent-distribution}
\end{figure}

\paragraph{\bf{Analysis.}}
As shown in \cref{fig:vae-compare-plot}, our video VAE substantially outperforms the Cosmos and Mochi tokenizers.
\cref{fig:vae-qualitative} and \cref{tab:vae_quan_comp} present qualitative and quantitative comparisons on videos featuring fast human motion and self-occlusion. 
Our model more effectively preserves structural and high-frequency details while introducing less visual distortion compared to the baselines.
To comprehensively evaluate the suitability of the latent space for generation, we visualize the latent distributions on the evaluation datasets in \cref{fig:vae-latent-distribution}. Our model produces a structured latent distribution that more closely approximates a Gaussian distribution than some other latent spaces, indicating that our latent space might be easier to be learned by a diffusion model than others.

Additional qualitative examples are included in the supplement. 

\begin{figure}[t]
\includegraphics[width=0.9\linewidth]{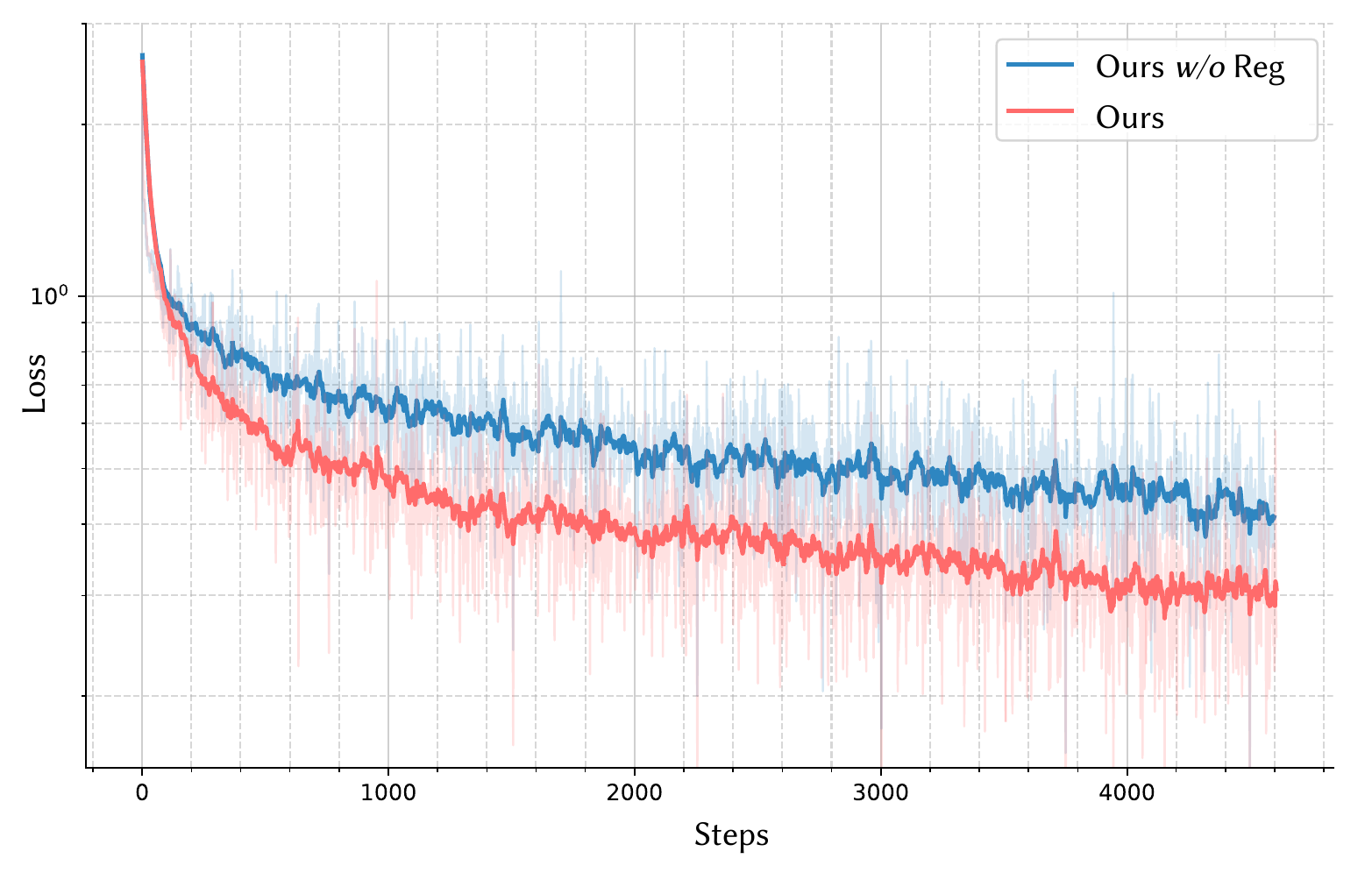}
  \caption{\textbf{Ablation study of latent regularization}. We compare training loss curves of diffusion transformers using latents from VAEs trained with and without regularization.}
  \label{fig:vae-ablation-study}
\end{figure}

\paragraph{\bf{Ablation on VAE regularization}} 
To evaluate our regularization term, we train diffusion transformers using latents from two of our VAE variants -- with and without regularization. As shown in Fig.~\ref{fig:vae-ablation-study}, the training loss reveals that regularized VAE produces more structured latent distributions that facilitate better and faster diffusion model training.

\section{Attention for 4D Human Video Generation}

Attention~\cite{vaswani2017attention} is widely recognized as a fundamental mechanism for capturing spatial relationships in sequences or images.
However, standard attention operations require comparing pairwise correlations in the data, making them inefficient when transitioning from 2D images to 3D or 4D generation tasks.
In this section, we first review the basics of self-attention and cross-attention mechanisms. We then introduce a novel \textit{interspatial} attention formulation that uses correspondences 
between image frames and parametric template meshes, thereby enabling efficient 4D human video generation.

\subsection{Basic Attention Mechanisms} 
\label{sec:attention:review}

\paragraph{\bf{Self-Attention.}}
Attention enables networks to learn feature relationships through weighted importance scores. 
The basic self-attention operation is formulated as:
\begin{align}
Q = X W_q , \quad K &= X W_k , \quad V =  X W_v, \\
\mathrm{\textsc{Attention}}(Q, K, V) &= \text{softmax}\left(\frac{QK^T}{\sqrt{d}}\right)V, 
\end{align}
where $Q$, $K$ and $V$ denotes query, key, and value matrices, respectively. $W_q \in \mathbb{R}^{d \times d_x}$, $W_k \in \mathbb{R}^{d \times d_x} $ and $W_v \in \mathbb{R}^{d \times d_x}$ are the learned projection matrix of input feature $X$, where $d$ is the learned feature dimension and $d_x$ is the input feature dimension.

\paragraph{\bf{Cross-Attention.}}
The attention mechanism can be extended to relate different feature spaces, thereby modeling inter-modality relationships.
In this case, $Q$ comes from one domain while $K$, $V$ come from another. A common example is to relate text features $Y$ and image features $X$ in text-conditioned diffusion models~\cite{rombach2022high} as:
\begin{align}
Q_x = X W_q , \quad K_y &= Y W_k , \quad V_y =  Y W_v, \\
\mathrm{\textsc{CrossAttention}}(X, Y) &= \text{softmax}\left(\frac{Q_xK_y^T}{\sqrt{d}}\right)V_y. 
\end{align}
Although text-to-image generation is a popular application, cross-attention generalizes to other modalities beyond text and images.

\paragraph{\bf{Transformer Block Integration.}}
Attention modules are frequently combined with Layer Normalization (LayerNorm) and a Feed-Forward Network (FFN) to form transformer blocks, which serve as a fundamental component in many diffusion models. Specifically, a transformer block $Y = \mathrm{\textsc{Transformer}}(X) $ comprises:
\begin{align}
\begin{aligned}
X &= X + \mathrm{\textsc{Attention}}(\mathrm{LayerNorm}(X)), \\
Y &= X + \mathrm{FFN}(\mathrm{LayerNorm}(X)),
\end{aligned}
\label{eq:transformer}
\end{align}
where $X$ represents the input features, and the attention operation could be either self-attention within the same modality or cross-attention between different modalities as described above.

\subsection{Interspatial Attention}
\label{sec:attention:ISA}
\label{sec:interspatial_attention}

Basic attention mechanisms learn correlations between different parts of an image or video by considering all other parts as equally viable candidates,
thus providing flexibility but suffering from inefficiency for 3D or 4D generation tasks.
In the context of 4D human video generation, the question arises: how can we effectively identify and attend to corresponding features across video frames without resorting to exhaustive comparisons between all features?
Our interspatial attention (ISA) mechanism builds on an intuitive insight: when generating 4D human videos conditioned on SMPL poses, 
the SMPL template provides rough correspondences across frames.
We leverage these correspondences to design an attention mechanism for 4D human video generation, which informs the network where to look for relevant correspondences.

\begin{figure}[t!]
  \includegraphics[width=0.8\linewidth]{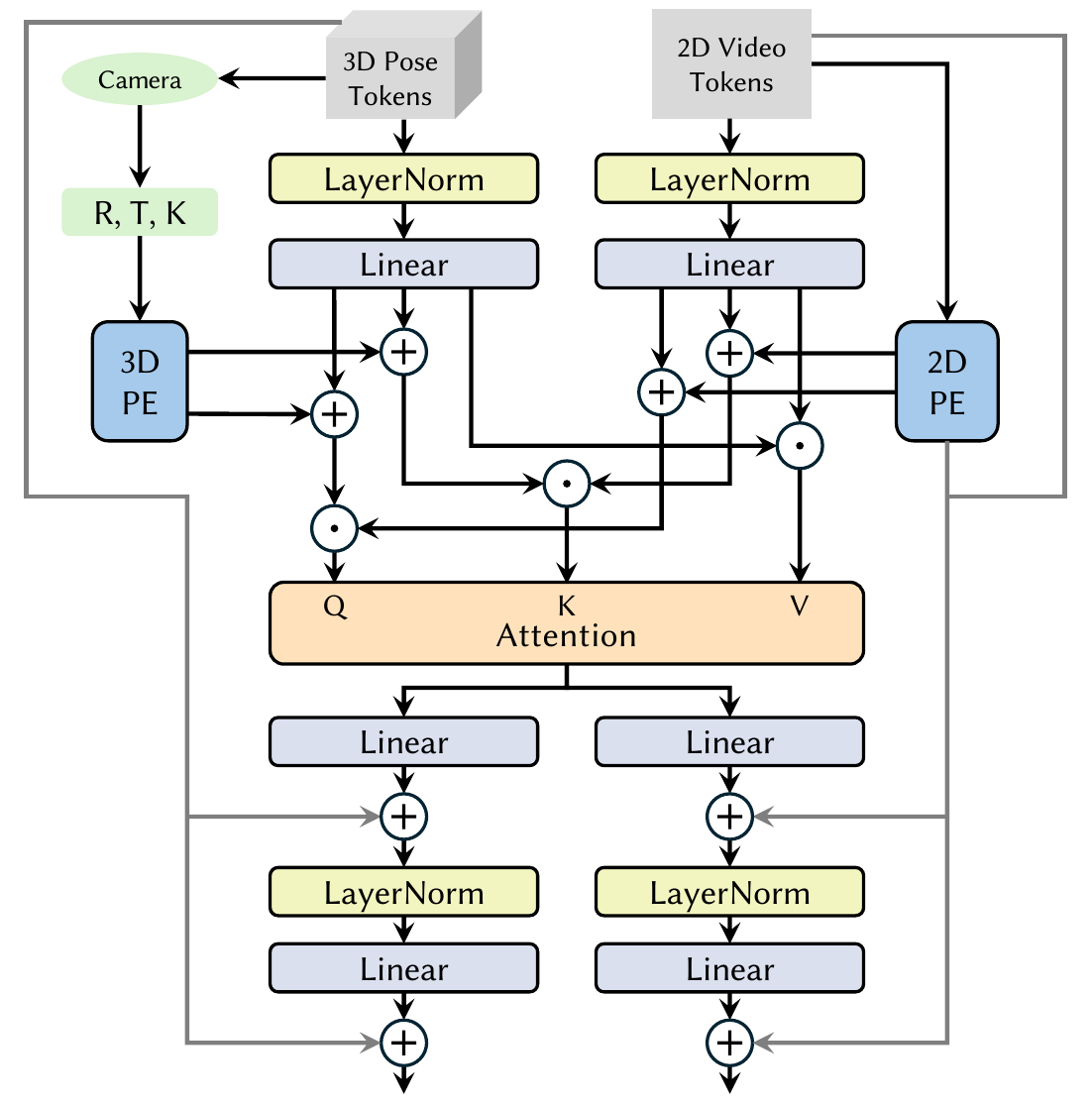}
  \caption{\textbf{Symmetric Interspatial Attention Block}. The attention block is a symmetric operation on 3D SMPL tokens and 2D video tokens. Concatenation is indicated by $\bigodot$ and element-wise addition by $+$.}
\label{fig:interspatial_attention}
\end{figure}

ISA implements this intuition in an efficient manner. Inspired by cross attention, ISA is an attention mechanism tailored for 4D human video generation that includes a carefully designed relative interspatial position encoding to provide the correct 4D geometric cues, enabling networks to capture the inherent 3D--2D relationships, as illustrated in Fig.~\ref{fig:interspatial_attention}.

\paragraph{\bf{Interspatial Attention without Positional Encoding.}}

To incorporate correspondences between the same parts of a digital human in different frames, we leverage the deformable 3D SMPL representation in combination with a cross-attention mechanism. This enables direct interaction between 3D human pose and 2D video features.

Specifically, we first sample a set of points on the surface of a SMPL mesh and construct a point sequence in global coordinate frame $\mathcal{G} = \{\mathbf{G}_i\}_{i=1}^{1+T}$, which we then convert into 3D tokens $\mathcal{Y} = \{\mathbf{Y}_i\}_{i=1}^{1+t}$ by a shallow MLP encoder $\mathrm{F_{mlp}}(\cdot)$ using the sinusoidal position encoding~\citep{vaswani2017attention} $\mathrm{PE}(\cdot)$:
\begin{equation}
\label{eqn:3d_tokens}
     \mathbf{Y}_{i}=\mathrm{F_{mlp}}(\mathrm{PE}(\mathbf{G}_i)).
\end{equation}

The 2D latents are then transformed from raw videos with our video VAE encoder: $\mathcal{Z} = \{\mathbf{z}_i\}_{i=1}^{1+t} = \mathrm{E}(\mathcal{V}=\{\mathbf{v}_i\}_{i=1}^{1+T})$ with a temporal downsample factor of $f_t=T/t$. Because the latents are downsampled along the temporal dimension, we conditioned a single latent $\mathbf{z}_{j}$ with the corresponding SMPL poses $\mathcal{Y}_{j} = \{\mathbf{Y}_i\}_{i=1+f_t\times (j-1)}^{1+f_t\times j}$ using cross attention: 
%
\begin{align}
\mathbf{z}'_j = \mathrm{\textsc{CrossAttention}}(\mathrm{Q}(\mathbf{z}_j), \mathrm{K}(\mathcal{Y}_{j}), \mathrm{V}(\mathcal{Y}_{j})),
\end{align}
where $\mathrm{Q}(\cdot)$, $\mathrm{K}(\cdot)$ and $\mathrm{V}(\cdot)$ are flattened learnable linear projection.
However, we find this simple cross attention leads to poor training convergence and fails to achieve accurate 3D pose conditioning (see e.g. Fig.~\ref{fig:ispe-training-curve}).
The network is required to infer geometric correspondences between the 2D video data and the 3D SMPL poses in the absence of explicit guidance, leading to suboptimal results.

\paragraph{\bf{Interspatial Positional Encoding (ISPE)}}
Inspired by implicit coordinate networks~\cite{deepsdf, nerf, occupancy}, we introduce ISPE to explicitly guide the network in building 3D--2D relationships. ISPE aims to model the spatial correspondence between 3D SMPL tokens and 2D video tokens by transforming their coordinates into a unified coordinate system using the known camera parameters.
Specifically, we project the coordinates of 3D SMPL tokens $\mathbf{g} = \left(x,y,z,w=1\right)$ to normalized device coordinate (NDC) space using the modelview-projection matrix $\mathbf{M}$:
\begin{equation}
\begin{split}
    \mathbf{g}_{clip} & = \left[x_{clip}, y_{clip}, z_{clip}, w_{clip} \right]^T = \mathbf{M}\mathbf{g}\,, \\
    \mathbf{g}_{ndc} & = \left[ \frac{x_{clip}}{w_{clip}}, \frac{y_{clip}}{w_{clip}}, \frac{z_{clip}}{w_{clip}} \right]^T.
\end{split}
\end{equation}
For 2D video tokens, we project their coordinates in (latent) pixel space $\mathbf{s} = [s_x, s_y]^T$ onto a 3D plane with zero depth in NDC space:
\begin{equation}
    \mathbf{s}_{ndc} = (2 \, s_x / w - 1, 2\, s_y / h - 1, 0),
\end{equation}
where $s_x$ and $s_y$ are the image coordinates at the latents, and $w,h$ denote the latent width and height.
After obtaining the coordinates in this unified (NDC) space, we apply a sinusoidal positional encoding $\mathrm{PE}(\cdot)$ to compute the ISPE.
We then incorporate ISPE into our proposed interspatial attention by adding it to both token:
\begin{align}
    \mathbf{z}'_{j} = \mathrm{\textsc{ISAttention}}&(\mathrm{Q}(\mathbf{z}_j + \mathrm{PE}(\mathbf{s}_{ndc})), \nonumber\\ 
    &\mathrm{K}(\mathcal{Y}_{j} + \mathrm{PE}(\mathbf{g}_{ndc})), \mathrm{V}(\mathcal{Y}_{j} + \mathrm{PE}(\mathbf{g}_{ndc}))).
\end{align}

By encoding features in a unified coordinate system, ISPE provides explicit geometric guidance for the attention mechanism. This spatial awareness helps establish effective 3D--2D correspondences during feature interaction, improving the quality of 3D conditioning.

\begin{figure*}
  \includegraphics[width=\linewidth]{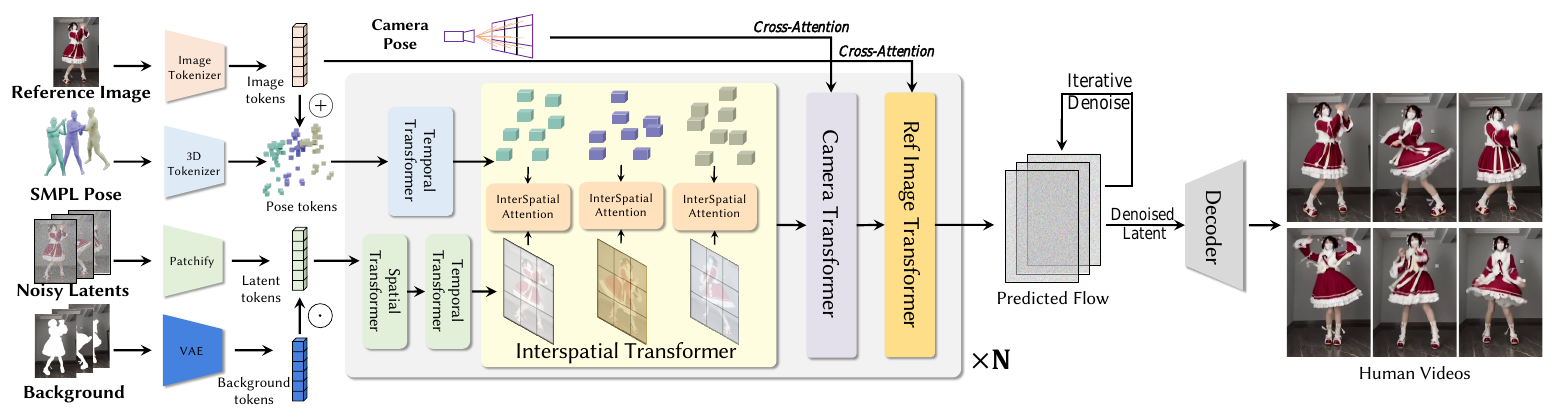}
  \caption{\textbf{ISA-DiT pipeline}. Overview of our diffusion transformer architecture for 4D human generation taking the reference image, SMPL condition, camera poses, and background videos as input.
  Our framework starts by tokenizing 3D SMPL conditions.
  In parallel, 2D video tokens (i.e., ``noisy latents'') are optionally composited with background elements and processed through a cascade of disentangled spatial and temporal transformer blocks, enabling efficient modeling of spatio-temporal relationships. These tokens then seamlessly interact with pose tokens via our Interspatial Transformer Block, facilitating effective 3D-aware conditioning. The generated features are further enhanced through Plücker camera embeddings for precise view control and interact with reference image features through cross attention to ensure consistent identity preservation. The entire framework is optimized using a flow-based diffusion formulation, enabling high-quality 4D human generation with controllable pose, viewpoint, and identity.}
  \label{fig:pipeline}
\end{figure*}

\paragraph{\bf{Symmetric Interspatial Attention (ISA)}}
Unlike previous approaches that only condition the video generation model using 2D projections of the SMPL template from a fixed viewpoint~\cite{xu2023magicanimate, hu2023animateanyone, zhu2024champ}, we propose a symmetric interspatial attention mechanism that enables bidirectional information flow between 3D and 2D spaces, inspired by the mm-DiT block in SD3~\cite{sd3}.  Specifically, we utilize 3D and 2D token features as queries and values respectively, with the ISPE guiding the attention:
\begin{align}
    \mathcal{Y}'_j = \mathrm{\textsc{ISAttention}}&(\mathrm{Q}(\mathcal{Y}_{j} + \mathrm{PE}(\mathbf{g}_{ndc})), \nonumber\\ 
    &\mathrm{K}(\mathbf{z}_j + \mathrm{PE}(\mathbf{s}_{ndc}))), \mathrm{V}(\mathbf{z}_j + \mathrm{PE}(\mathbf{s}_{ndc}))), \label{equ:ISAY}\\
    \mathbf{z}'_{j} = \mathrm{\textsc{ISAttention}}&(\mathrm{Q}(\mathbf{z}_j + \mathrm{PE}(\mathbf{s}_{ndc})), \nonumber\\ 
    &\mathrm{K}(\mathcal{Y}_{j} + \mathrm{PE}(\mathbf{g}_{ndc})), \mathrm{V}(\mathcal{Y}_{j} + \mathrm{PE}(\mathbf{g}_{ndc}))), \label{equ:ISAz}
\end{align}

In this way, our approach implicitly performs simultaneous rendering (3D-to-2D) and reconstruction (2D-to-3D) by allowing features to interact in both directions. This improved feature interaction results in more effective conditioning of 3D structural information, facilitating consistent and high-quality 4D human video generation. As shown in \cref{fig:interspatial_attention}, we then integrate the symmetric ISA with LayerNorm~\citep{lei2016layer} and a Feedforward Network (FFN)~\citep{vaswani2017attention} to form a Symmetric Interspatial Transformer Block, similar to \cref{eq:transformer}. This block serves as a crucial component in our video diffusion transformer architecture.

\subsection{Interspatial Diffusion Transformer (ISA-DiT)}
\label{sec:attention:ISADiT}

Based on our video VAE and the ISA attention, we now discuss how to integrate it into a diffusion transformer architecture for human video generation that effectively bridges 3D structural information and 2D video features. The core change to a conventional DiT architecture is the addition of parallel symmetric branches: a 3D branch for learning SMPL features and a 2D branch for video features. These branches are interconnected through our ISA block. Our framework, illustrated in Fig.~\ref{fig:pipeline}, uses a single input image as conditioning information and simultaneously injects the included human identity into both 3D and 2D branches for identity consistency across generated frames. Furthermore, we introduce a switchable background conditioning module, which enables flexible composition between human videos and various background settings. We discuss the unique components of our architecture in the following.

\paragraph{\bf{Symmetric Diffusion Branch.}}
Our framework employs a symmetric diffusion architecture comprising specialized transformer modules (Fig.~\ref{fig:pipeline}). Building upon SD3's architecture, we extend it for video generation by incorporating temporal transformer blocks between existing 2D image transformer blocks. In this enhanced architecture, 2D video tokens $\mathbf{z}$ are first processed through a spatial transformer block:
\begin{equation}
\mathbf{z}_s = \mathrm{\textsc{SpatialTransformer}}(\mathbf{z}),
\end{equation}
followed by a temporal transformer that establishes frame-wise temporal correlations:
\begin{equation}
\mathbf{z}_{st} = \mathrm{\textsc{TemporalTransformer}}(\mathbf{z}_s).
\end{equation}
For effective 3D SMPL conditioning, we encode sampled SMPL points into 3D tokens as described in \cref{eqn:3d_tokens}. The 3D SMPL tokens $\mathbf{Y}$ are processed through a temporal transformer block to establish temporal continuity across consecutive SMPL representations:
\begin{equation}
\mathbf{Y}_{t} = \mathrm{\textsc{TemporalTransformer}}(\mathbf{Y}).
\end{equation}
These 3D tokens are then processed by the symmetric ISA transformer block, which bridges the gap between 3D SMPL poses and 2D videos, enabling seamless interaction:
\begin{equation}
\mathbf{Y}_t', \mathbf{z}_{st}' = \mathrm{\textsc{ISATransformer}}(\mathbf{Y}_{t}, \mathbf{z}_{st}).
\end{equation}
Finally, the learned 2D video features interact with the camera pose and reference image through cross-attention blocks, which we describe in detail in the following.

\paragraph{\bf{Identity Conditioning Module.}}
To ensure identity consistency across diverse views and temporal frames, we propose a novel identity injection strategy that simultaneously incorporates human identity information into both 3D SMPL and 2D video features. \revise{Specifically, our identity condition module has two sub-modules. }

For the 3D SMPL branch, we first extract the latent feature $\mathbf{z}_{ref}$ from reference image $\mathbf{I}_{ref}$ with the VideoVAE.
\revise{We then estimate the SMPL models of the reference image} and perform pixel-aligned feature propagation onto 3D SMPL tokens:
\begin{equation}
\mathbf{Y} = \mathbf{Y} + \mathrm{\textsc{GridSample}}(\mathbf{z}_{ref}, \pi_y(\mathbf{Y})),
\end{equation}
\revise{where $\pi_y(\mathbf{Y})$ are the projected 2D reference image coordinates for the 3D SMPL tokens}. This 3D propagation enhances the temporal consistency of human identity.

For 2D video features, we inject the reference image through both local concatenation and global cross-attention mechanisms:
\begin{align}
\mathbf{z}_{l} &= \mathrm{\textsc{Concat}}([\mathbf{z}, \mathbf{z}_{ref}]), \\
\mathbf{z}_{g} &= \mathrm{\textsc{CrossAttention}}(\mathbf{z}, \mathrm{\textsc{CLIP}}(\mathbf{I}_{ref})), 
\end{align}
where the concatenation operation preserves fine-grained identity details, and the CLIP embedding in cross-attention ensures global identity consistency.

\paragraph{\bf{Camera Conditioning Module.}}
To achieve precise camera view control, we parameterize the camera poses into Plücker coordinates for detailed geometric modeling following prior works~\cite{he2024cameractrl}.
We first encode the rotation and translation of camera parameters into  Plücker images $\mathbf{c}$. Considering the temporal downsampling of our videos, we concatenate multiple camera embeddings corresponding to the same latent frame across channels:
\begin{equation}
\mathbf{c}_{latent} = \mathrm{\textsc{Concat}}([\mathbf{c}_1, ..., \mathbf{c}_k]),
\end{equation}
and the camera condition is then injected via cross-attention:
\begin{equation}
\mathbf{z}_{cam} = \mathrm{\textsc{CrossAttention}}(\mathbf{z}, \mathbf{c}_{latent}).
\end{equation}

\paragraph{\bf{Background Conditioning Module.}}
Our framework enables flexible background composition through a conditional injection mechanism. The background videos $\mathbf{v}_{bg}$ are first encoded through our videoVAE encoder into a set of latents:
$\mathbf{z}_{bg} = \mathrm{E}(\mathbf{v}_{bg}).$
The background features are then integrated with the main video latents through concatenation: $\mathbf{z}_{final} = \mathrm{\textsc{Concat}}([\mathbf{z}, \mathbf{z}_{bg}])$. For scenarios without background composition, we utilize a zero latent: $\mathbf{z}_{final} = \mathrm{\textsc{Concat}}([\mathbf{z}, \mathbf{0}])$.

\paragraph{\bf{Diffusion Formulation.}}
Inspired by SD3~\cite{sd3}, our framework adopts flow matching for the diffusion process. Given a timestep $t$, we perturb the original video $x_0$ following: $x_t = (1-t)x_0 + t\epsilon, \,\,\epsilon \sim \mathcal{N}(0, \mathbf{I}).$ The network is trained to predict the flow field $\mathbf{v} = x_0 - \epsilon$, offering improved stability and training efficiency compared to conventional diffusion approaches.

\paragraph{\bf{Implementation Details.}} Additional details on network architecture design, hyperparameter selection, and training are included in the supplement. We will release code and pretrained model.

\begin{table*}[t]
\centering
 \caption{\textbf{Quantitative comparison of generated videos}. We compare our method with state-of-the-art baselines AnimateAnyone~\cite{hu2023animateanyone}, Champ~\cite{zhu2024champ}, MusePose~\cite{musepose}, Animate-X~\cite{tan2024animate}, and Human4DiT~\cite{shao2024human4dit} using multiple metrics (PSNR, SSIM, LPIPS, and FVD). Specifically, we evaluate three scenarios: videos with a static background (``Video''), with camera movement (``Camera''), and with background mask applied (``Mask''). Our approach achieves superior quality across all metrics and all scenarios. }
\resizebox{\textwidth}{!}{
\begin{tabular}{lcccccccccccc}
\toprule
\multirow{2}{*}{Method} & \multicolumn{3}{c}{PSNR $\uparrow$} & \multicolumn{3}{c}{SSIM $\uparrow$} & \multicolumn{3}{c}{LPIPS $\downarrow$} & \multicolumn{3}{c}{FVD $\downarrow$} \\
\cmidrule(lr){2-4} \cmidrule(lr){5-7} \cmidrule(lr){8-10} \cmidrule(lr){11-13}
& Video & Camera & Mask & Video & Camera & Mask & Video & Camera & Mask & Video & Camera & Mask \\
\midrule
AnimateAnyone & 19.39 & 18.87 & 26.99 & 0.757 & 0.656 & 0.953 & 0.211 & 0.234 & 0.058 & 693.9 & 935.9 & 234.7 \\
Champ & 21.72 & 20.04 & 26.34 & 0.819 & 0.712 & 0.954 & 0.126 & 0.204 & 0.041 & 466.8 & 872.3 & 181.7 \\
MusePose & 22.19 & 20.75 & 23.80 & 0.830 & 0.708 & 0.940 & 0.119 & 0.195 & 0.048 & 481.0 & 1135 & 264.4 \\
Animate-X & 23.03 & 21.67 & 29.84 & 0.839 & 0.719 & 0.965 & 0.116 & 0.163 & 0.039 & 285.3 & 608.4 & 147.7 \\
Human4DiT & 24.71 & 22.24 & 27.68 & 0.889 & 0.767 & 0.957 & 0.109 & 0.213 & 0.031 & 388.2 & 623.9 & 162.3 \\
ISA-DiT (ours) & \textbf{28.34} & \textbf{27.78} & \textbf{32.06} & \textbf{0.931} & \textbf{0.855} & \textbf{0.976} & \textbf{0.049} & \textbf{0.071} & \textbf{0.014} & \textbf{143.6} & \textbf{227.9} & \textbf{81.3} \\
\bottomrule
\end{tabular}}
  
   \label{tab:dview_comp}
\end{table*}

\begin{figure*}
\includegraphics[width=\textwidth]{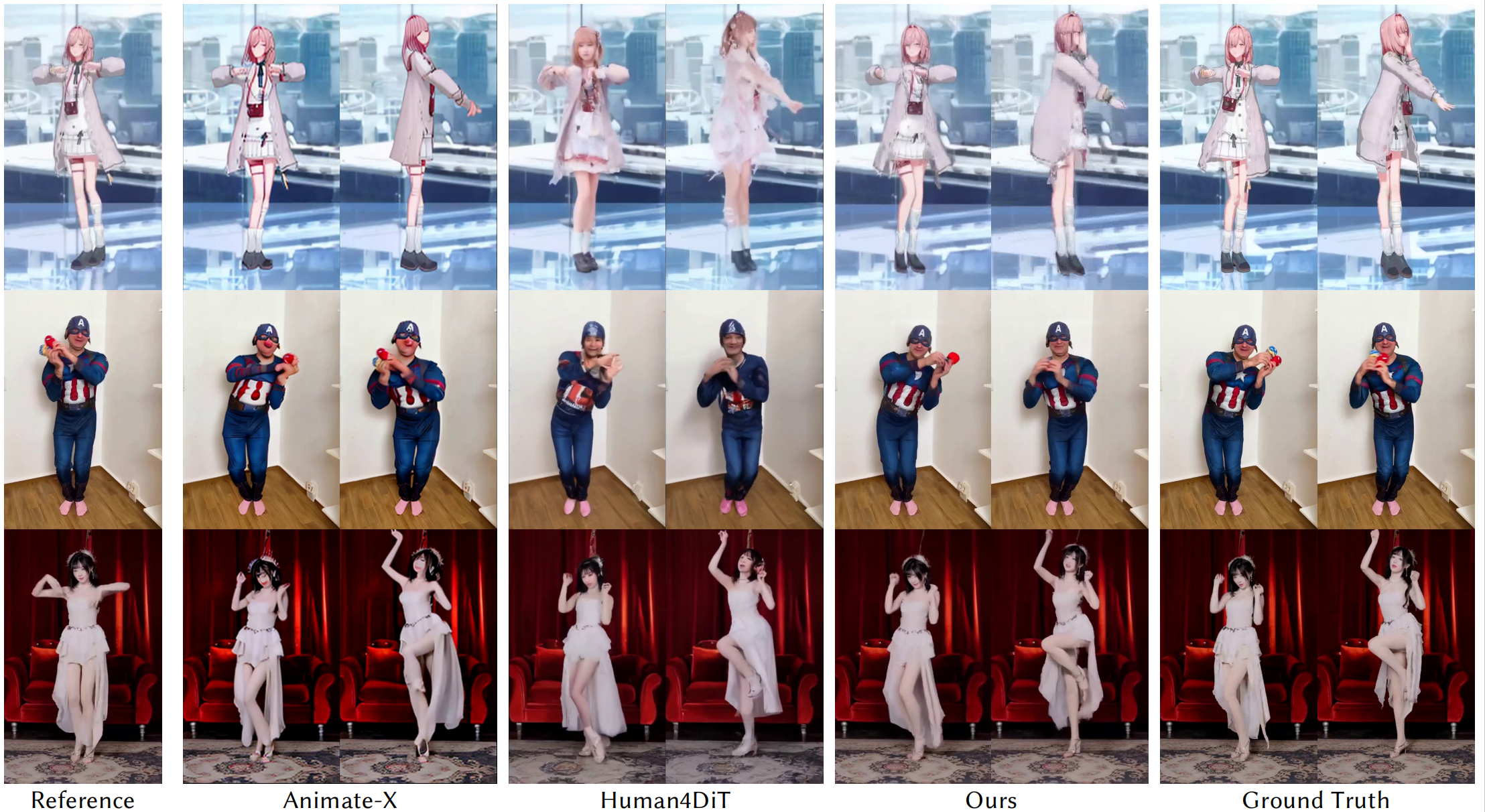}
  \caption{\textbf{Qualitative comparisons of generated videos.} We compare our approach with the best-performing baselines; each of these methods is conditioned on the reference image shown on the left.
  Our method achieves superior visual quality, particularly in capturing facial expressions, modeling clothing dynamics, and rendering natural hand--object interactions. 
  }
  \label{fig:video-gen-comparisons}
\end{figure*}

\section{Experiments}

\subsection{Data}

For \textit{VAE traning}, we utilize the Kinetics-600~\cite{kay2017kinetics} and Human4DiT~\cite{shao2024human4dit} datasets. Additionally, we curate a custom evaluation dataset comprising 200 human videos featuring multi-person interactions, complex textures, and fast motion sequences.

For \textit{DiT training}, we curate a dataset comprising 1M real human videos and 100K synthetic videos rendered using the PointOdyssey pipeline \cite{zheng2023pointodyssey}. \revise{The real human videos are sourced from existing datasets including Human4DiT~\cite{shao2024human4dit}, Pexel, OpenVID-1M~\cite{nan2024openvid}, MiraData~\cite{ju2024miradata}, and Koala-36M~\cite{wang2024koala}}. The synthetic data is generated using 200 digital human models animated with motion sequences sampled from the CMU~\cite{cmuWEB} and AMASS~\cite{neureen2019amass} datasets. These animations are rendered in 1,000 different environment maps using procedurally generated camera trajectories, producing 100K videos at resolutions ranging from $512\times512$ to $1024\times1024$. Since the synthetic data are highly controlled as we can directly export ground truth SMPL poses, camera poses, and backgrounds from the rendering engine. For the real human videos, we obtain SMPL annotations using Humans-in-4D~\cite{goel2023humansIn4d} and segment the background on 1M videos using SAM2~\cite{ravi2024sam}. For camera condition learning, we only utilize the camera poses from the synthetic data. 

For \textit{DiT evaluation}, we construct several test sets to evaluate different aspects of the model: 
\begin{itemize}
    \item ``Video'' Dataset: We collect 100 monocular human videos with static cameras, focusing on pure human motion without camera movement. This dataset serves to evaluate our method's performance in human animation generation.
    \item ``Camera'' Dataset: This subset consists of 100 videos featuring both human motion and camera movement. It is designed to assess our method's ability to generate 4D human scene videos with dynamic camera trajectories.
    \item ``Mask'' Dataset: We create a dedicated dataset of 100 videos with camera motion where backgrounds are masked out. This dataset evaluates the generated performance for the digital human(s) without considering the complex backgrounds.
\end{itemize}
These carefully curated datasets enable comprehensive evaluation of our model across three key aspects: human motion synthesis under static views, 4D human--scene generation with moving cameras, and isolated human motion generation with masked backgrounds.

\subsection{Settings}

\paragraph{\bf{Baselines.}}
We compare our method with several state-of-the-art approaches for human video generation, including AnimateAnyone~\cite{hu2023animateanyone}, CHAMP~\cite{zhu2024champ}, MusePose~\cite{musepose}, Animate-X~\cite{tan2024animate}, and Human4DiT~\cite{shao2024human4dit}. Since the official implementation of AnimateAnyone is not publicly available, we utilize the PyTorch version implemented by Moore-AnimateAnyone\footnote{https://github.com/MooreThreads/Moore-AnimateAnyone} for our experiments.
For AnimateAnyone, MusePose, and Animate-X, we employ DWPose~\cite{yang2023effective} to extract human pose estimations from the input videos, generating skeleton graphs as conditional inputs. For CHAMP, following the official pipeline, we simultaneously estimate SMPL parameters and render the corresponding depth maps, normal maps, and semantic segmentation masks of the motion videos, along with DWPose skeleton graphs as conditional inputs. For Human4DiT, we estimate SMPL parameters to render SMPL normal maps and use DPVO~\cite{teed2023deep} to estimate camera parameters from the motion videos as conditional inputs.
Given that all these methods employ image-based VAE architectures with limited temporal inference windows, we adopt a sliding window approach during inference with a window size of 24 frames and an overlap of 8 frames between consecutive windows. For fair comparison, all methods use 30 sampling steps with the DDIM~\cite{ddim} scheduler during inference.

\revise{
\paragraph{\bf{SOTA image-to-video models.}}
To evaluate the generative ability of our model for human-centric videos, we conduct comprehensive comparisons with SOTA image-to-video models, including Cosmos~\cite{agarwal2025cosmos}, Hunyuan~\cite{kong2024hunyuanvideo} and Wan~\cite{wang2025wan} using the VBench~\cite{huang2023vbench} benchmark suite.
}

\paragraph{\bf{Evaluation Metrics.}}
To evaluate our pose-driven image-to-video generation model, we employ both frame-level and video-level metrics. For individual frames, we assess generation quality using standard metrics: PSNR, SSIM, and LPIPS~\cite{lpips}. For video-level evaluation, we measure generation quality using Fréchet Video Distance (FVD)~\cite{unterthiner2019fvd}.

\revise{
\paragraph{\bf{CFG Scales.}}
We observe that classifier-free guidance (CFG) plays a crucial role in video generation quality. At CFG=1, generated videos exhibit noticeable blurriness, motion blur, and lack fine details. Conversely, high CFG values produce overly sharp videos with texture artifacts. In our experiments, we employ CFG=2 while comparison methods are evaluated using their default CFG settings.
}

\begin{figure*}
\includegraphics[width=\linewidth]{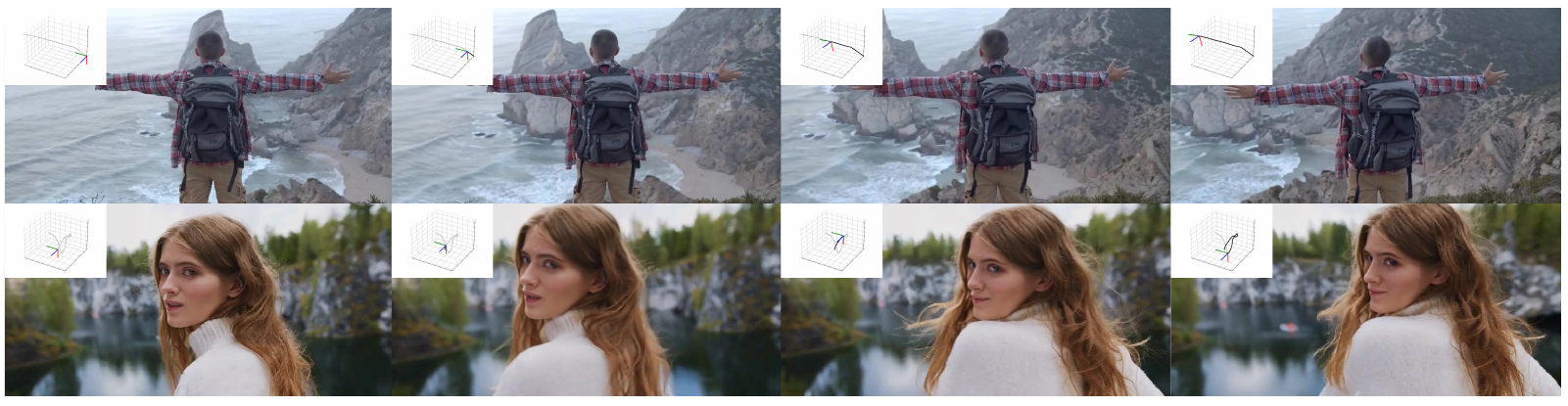}
  \caption{ \textbf{Generating videos with controllable camera trajectories.} Our model can generate high-quality human videos conditioned on specific camera trajectories (top left insets), effectively transforming video generation into a dynamic view-synthesis system for multi-view human generation. 
  }
  \label{fig:video-gen-camera-control}
\end{figure*}

\begin{figure*}
\includegraphics[width=\linewidth]{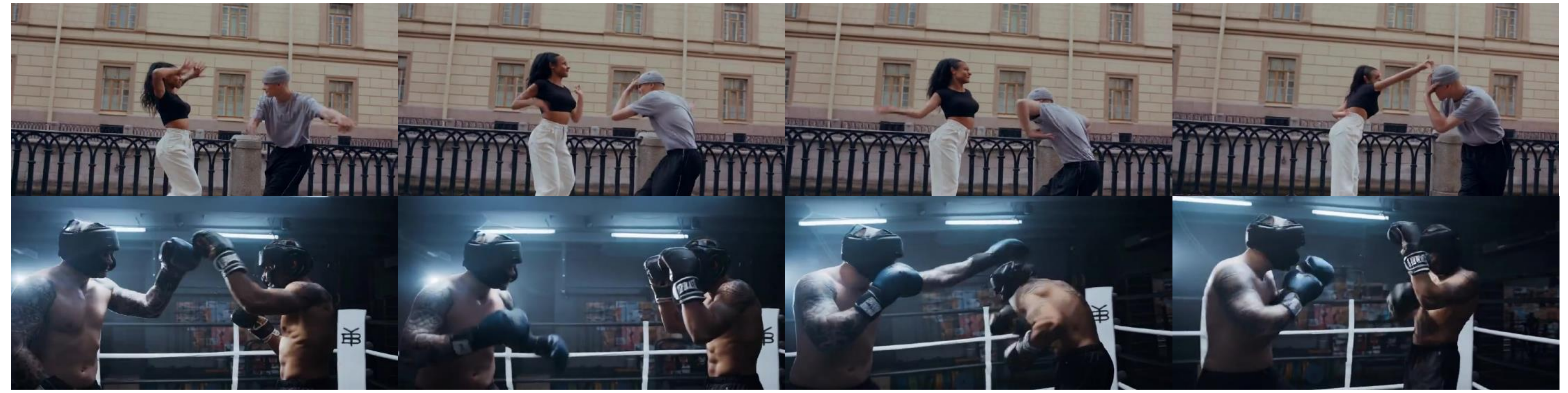}
  \caption{\textbf{Generating multiple characters.} 
Our method synthesizes multi-character videos featuring realistic interactions, such as dancing and boxing. 
  }
  \label{fig:video-gen-multiperson}
\end{figure*}

\begin{figure*}
\includegraphics[width=\textwidth]{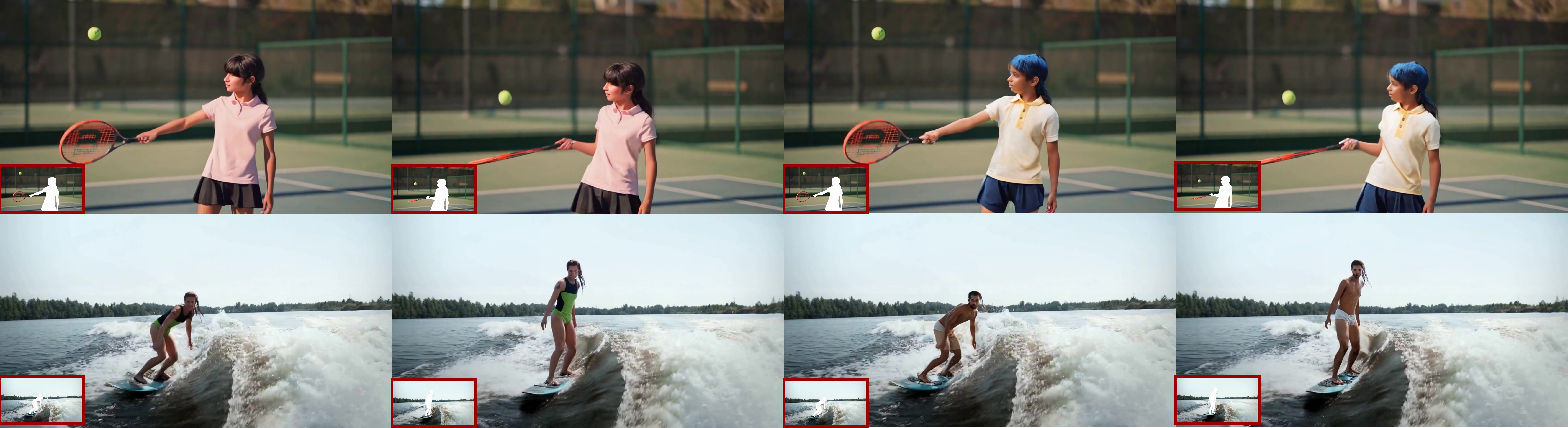}
  \caption{\textbf{Human video generation with controlled backgrounds.} Our method generates videos by compositing synthesized digital humans with background scenes, achieving consistent lighting and shadow effects based on background conditions. 
  }
  \label{fig:video-background-comp}
\end{figure*}

\begin{figure*}
\includegraphics[width=.95\textwidth]{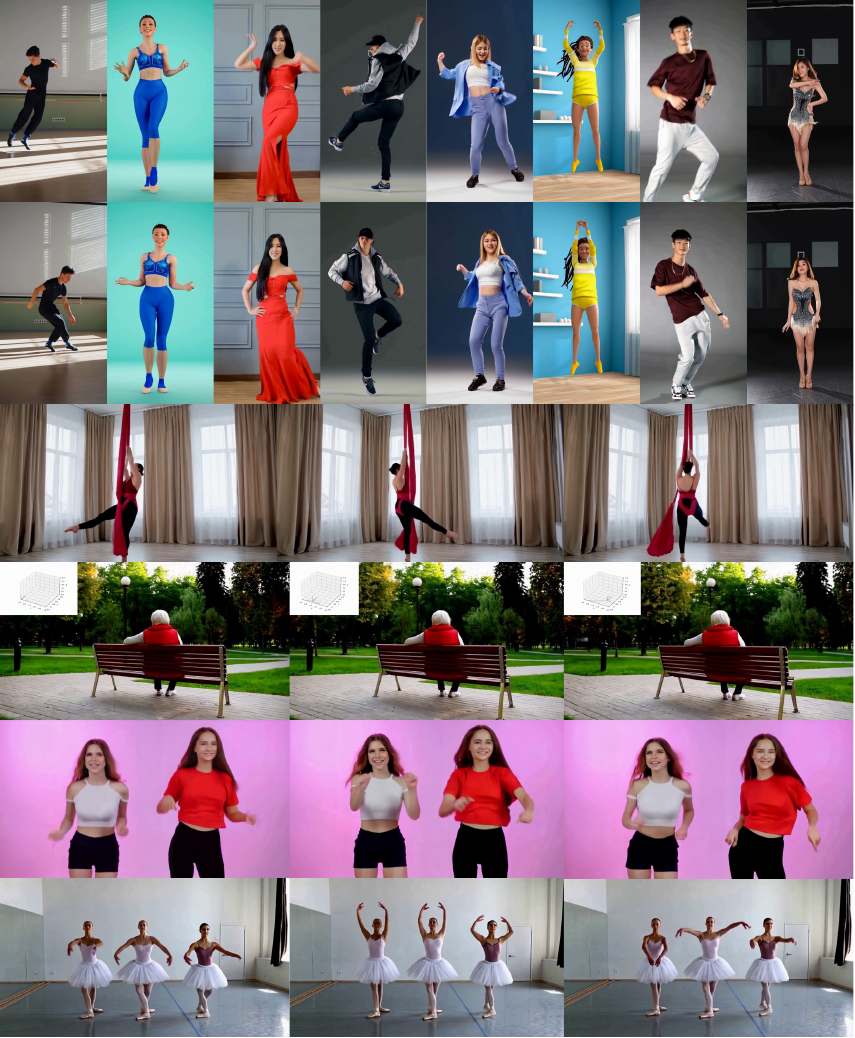}
  \caption{\textbf{Additional generated videos.} Our ISA-DiT framework generates high-quality videos across diverse domains, spanning upper-body portraits, full-body movements, anime character and multi-characters animations. 
  }
  \label{fig:video-gen-our-results}
\end{figure*}

\subsection{Evaluation}

\begin{table}[t!]
    \centering
    \caption{\revise{\textbf{Quantitative comparison based on VBench}. We compare our method with state-of-the-art image-to-video methods including Cosmos~\cite{cosmos}, Hunyuan~\cite{kong2024hunyuanvideo}, and WAN2.1~\cite{wang2025wan} using multiple metrics (Quality, Aesthetics, and Consistency).}}
    \begin{tabular}{c|c|c|c}
    \toprule
       Method  & Quality$\uparrow$ & Aesthetics$\uparrow$ & Consistency$\uparrow$ \\  \hline 
      Cosmos-I2V-14B   & 0.693 & 0.528 & 0.896 \\
      Hunyuan-I2V-14B & 0.705 & 0.546 & 0.904 \\
      WAN2.1-I2V-14B & \textbf{0.738} & \textbf{0.582} & 0.929 \\
      ISA-DiT(ours)-4B & 0.724 & 0.579 & \textbf{0.953} \\ 
      \bottomrule
    \end{tabular}
    \label{tab:vbench}
\end{table}

\paragraph{Quantitative Comparisons.} We perform comprehensive quantitative comparisons on the DiT evaluation datasets in \cref{tab:dview_comp}. 
Our ISA-DiT shows improvements over the baselines across key metrics, including PSNR, SSIM, LPIPS, and FVD. These quantitative results suggest that our model is better at generating detailed and consistent human videos. Our method performs particularly well in scenarios involving camera motion, demonstrating effective handling of both human animation and scene dynamics. 

\revise{Additionally, we utilize Vbench to evaluate and compare the generation capability of our model with current open-source image-to-video generation models, where we select the first frame of the videos in our proposed ``Video Dataset'' as the input image. As shown in \cref{tab:vbench}, our model achieves performance comparable to current open-source large video models, with significantly smaller model sizes and greater efficiency, demonstrating the effectiveness of our proposed ISA and model design.} 

\paragraph{Qualitative Comparisons.} 
We compare our method qualitatively against state-of-the-art approaches, with results shown in \cref{fig:video-gen-comparisons} and the supplemental video. Here, we focus on the best-performing baselines---Human4DiT and Animate-X; additional results are provided in the supplement. Our method demonstrates superior quality in human video generation, which is most noticeable in facial details, body structure, and dynamic motion. Our generated videos also show improved quality in natural movements, including realistic hair dynamics, clothing deformation, and hand--object interactions. These results validate our method's ability to effectively learn to generate complex dynamic features, leading to more coherent and realistic human videos.

\paragraph{Camera Control.}
We present generated human videos with diverse camera trajectories in \cref{fig:video-gen-camera-control} and the supplemental video, illustrating our method’s ability to jointly control human motion and background changes under camera movement. 
Leveraging our interspatial attention block, our model produces view‐consistent human–scene videos with a high level of multi-view consistency and dynamic camera control.

\paragraph{Multi-character Animation.}
Our method supports multiple digital humans in the same video, as shown in \cref{fig:video-gen-multiperson}. This capability stems from our identity control module and ISA mechanism, which flexibly maps between generated video content and SMPL conditions regardless of the number of input characters. Specifically, we first track and obtain the SMPL for each individual, then sample points from each person's SMPL to generate tokens which are concatenated and fed into the ISA block. Through this injection, we effectively maintain identity consistency and achieve superior spatiotemporal coherence.

\paragraph{Background Composition.} Our method also enables creative applications in human--background video compositing. As demonstrated in \cref{fig:video-background-comp}, our method is able to generate different characters using the same background video, with composite videos maintain consistency in lighting, shadows, and perspective between the generated human and the background environment. Similarly, we could generate the same character in front of different backgrounds (not shown).

\paragraph{Additional Results.}  To comprehensively showcase the capabilities of our method, we present an extensive collection of generation results in \cref{fig:video-gen-our-results} and the supplementary video. Our method handles a diverse range of scenarios, including facial animations, upper-body portrait videos, full-body animations, multi-person interactions, and anime character generation. The consistent performance across varied applications indicates the method's adaptability and generalization capabilities.

\begin{figure}
\includegraphics[width=\linewidth]{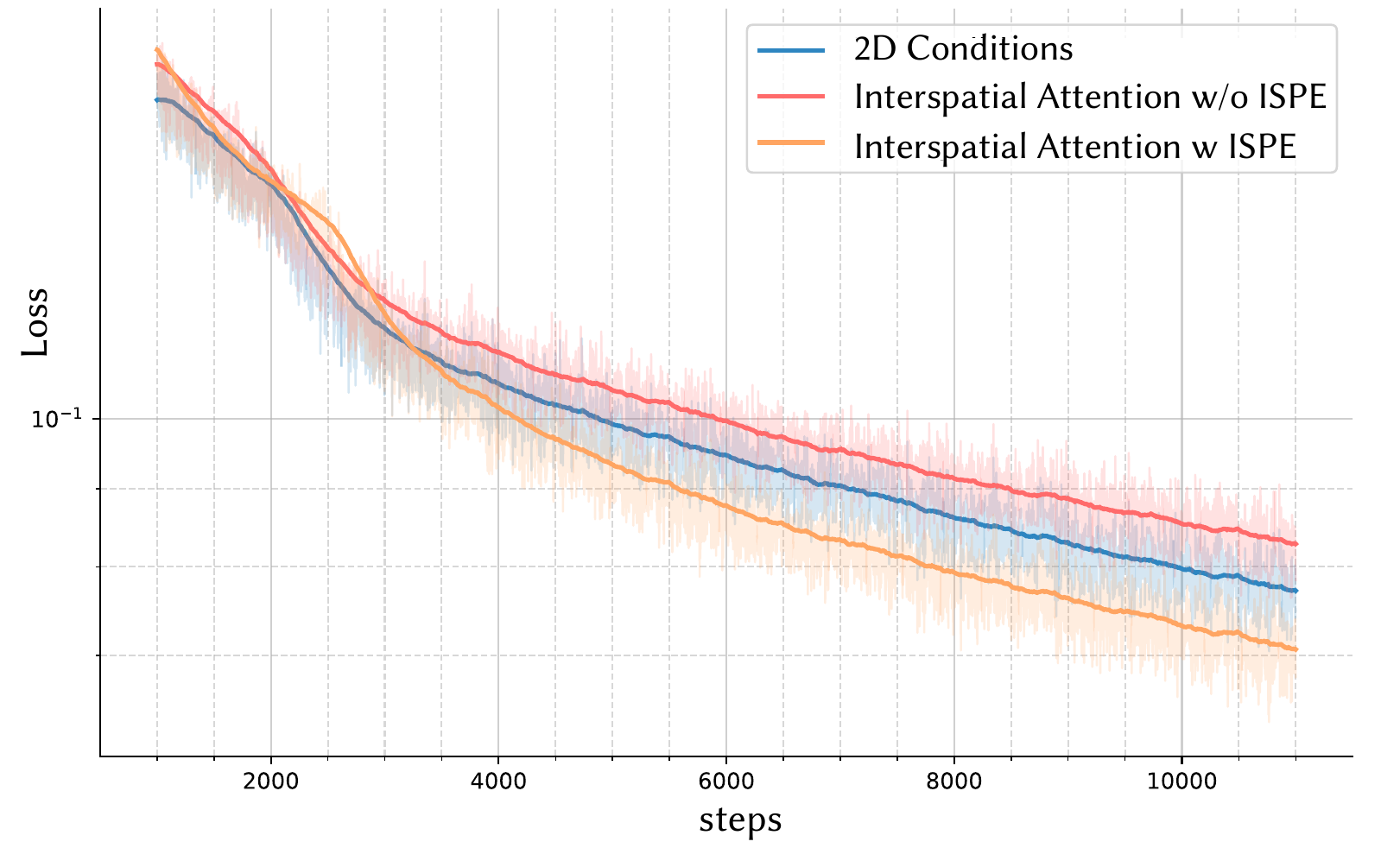}
  \caption{\textbf{Ablation of interspatial attention.} We compare validation loss curves for the same DiT architecture using three different conditioning mechanisms: a baseline that only uses the 2D SMPL normal maps for conditioning, ISA without interspatial positional encoding, and interspatial attention with positional encoding. The latter conditioning converges faster and to a lower loss value than the other options.}
  \label{fig:ispe-training-curve}
\end{figure}

\begin{figure}
    \centering
    \includegraphics[width=.9\linewidth]{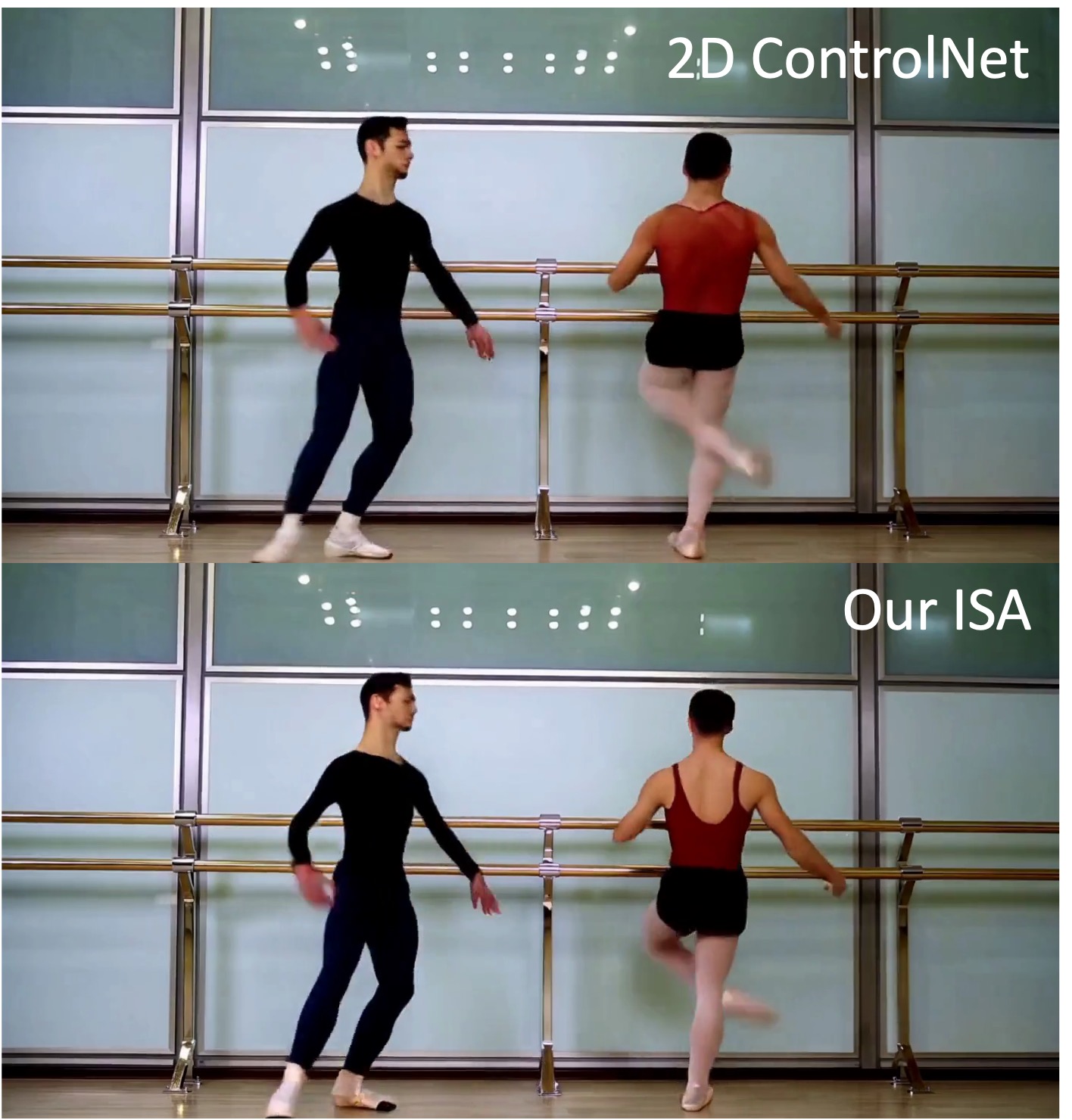}
    \caption{\revise{\textbf{Ablation of interspatial attention.} We show qualitative comparisons between 2D ControlNet and ISA. For complex 3D poses, the ControlNet produces implausible deformations while our ISA generates more natural and realistic human motions. }}
    \label{fig:ablation}
\end{figure}

\begin{table}[t!]
    \centering
    \caption{\revise{\textbf{Ablation Study.} We conduct an ablation study on our proposed ISA and different 3D template control signals. ``Ours (W/o ISPE)'' refers to using ISA without the proposed ISPE, while ``Ours (2D ControlNet)'' refers to not using ISA and instead employing 2D ControlNet for SMPL-based conditioning. Across different 3D templates, we compared the facial generation capabilities of both SMPL and FLAME models.}}
    \begin{tabular}{c|c|c|c|c}
    \toprule
       Method  & PSNR$\uparrow$ & SSIM$\uparrow$ & LPIPS$\downarrow$ & FVD$\downarrow$ \\  \hline 
       Ours (W/o ISPE) & 25.21 & 0.895 & 0.079 & 230.2 \\
        Ours (2D ControlNet) & 26.45 & 0.916 & 0.064 & 195.7 \\
      Ours   & \textbf{28.34} & \textbf{0.931} & \textbf{0.049} & \textbf{143.6} \\
       \hline
      Face (SMPL) & 30.42 & 0.955 & 0.039 & 112.4 \\
      Face (FLAME) & \textbf{31.05} & \textbf{0.972} & \textbf{0.034} & \textbf{101.9}\\
      \bottomrule
    \end{tabular}
    
    \label{tab:ablation}
\end{table}

\begin{figure}
    \centering
    \includegraphics[width=.9\linewidth]{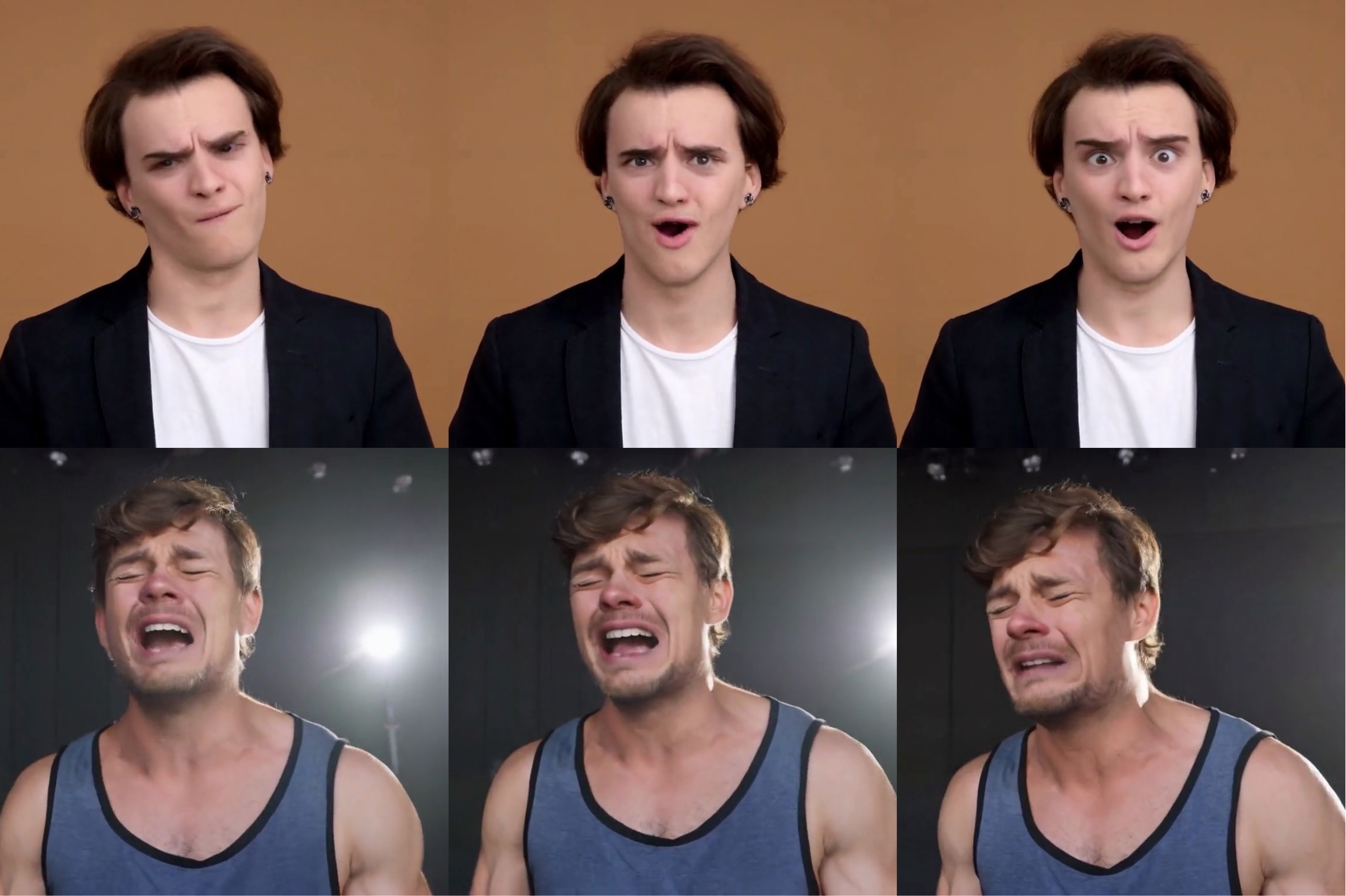}
    \caption{\revise{\textbf{ISA with 3D FLAME for expression generation.} ISA could be effectively integrated with more precise 3D face models like FLAME to achieve vivid facial generation.}}
    \label{fig:face}
\end{figure}

\subsection{Ablation Study}

\paragraph{ISA mechanism}
To validate the effectiveness of our ISA mechanism, we run an ablation to compare three variants of our DiT: (1) a baseline without ISA, where SMPL conditions are rendered directly into 2D normal maps, which are used as conditioning input; (2) ISA without position encoding; and (3) ISA with position encoding. \cref{fig:ispe-training-curve} shows the validation loss curves for each configuration. During early training, all variants converge quickly. As training progresses, the position-encoded ISA variant establishes stronger 3D--2D correlations leading to faster convergence and lower a validation loss compared to the other configurations. \revise{Additionally, \cref{fig:ablation} and \cref{tab:ablation} demonstrate qualitative and quantitative results. For videos involving fast movements and complex 3D poses, the ControlNet variant produces implausible deformations while our ISA generates more natural and realistic human motions.} This experiment clearly validates the effectiveness of our ISA design.

\revise{\paragraph{3D Template Model}
Additionally, we explore the use of an alternative 3D face model, i.e., FLAME~\cite{FLAME:SiggraphAsia2017}, to enhance ISA's facial modeling. We evaluate it on 100 face-centric videos using the FLAME model as conditions. \cref{tab:ablation} shows improvements across all metrics from SMPL to FLAME. With 3D FLAME, ISA can generate vivid human expressions as shown in \cref{fig:face}. This demonstrates ISA could be effectively integrated with more precise 3D face models to achieve superior facial generation.}

\subsection{Scaling.}

We perform additional experiments that validate the effectiveness of ISA when scaling the DiT architecture in the supplement.

\begin{figure}
    \centering
    \includegraphics[width=\linewidth]{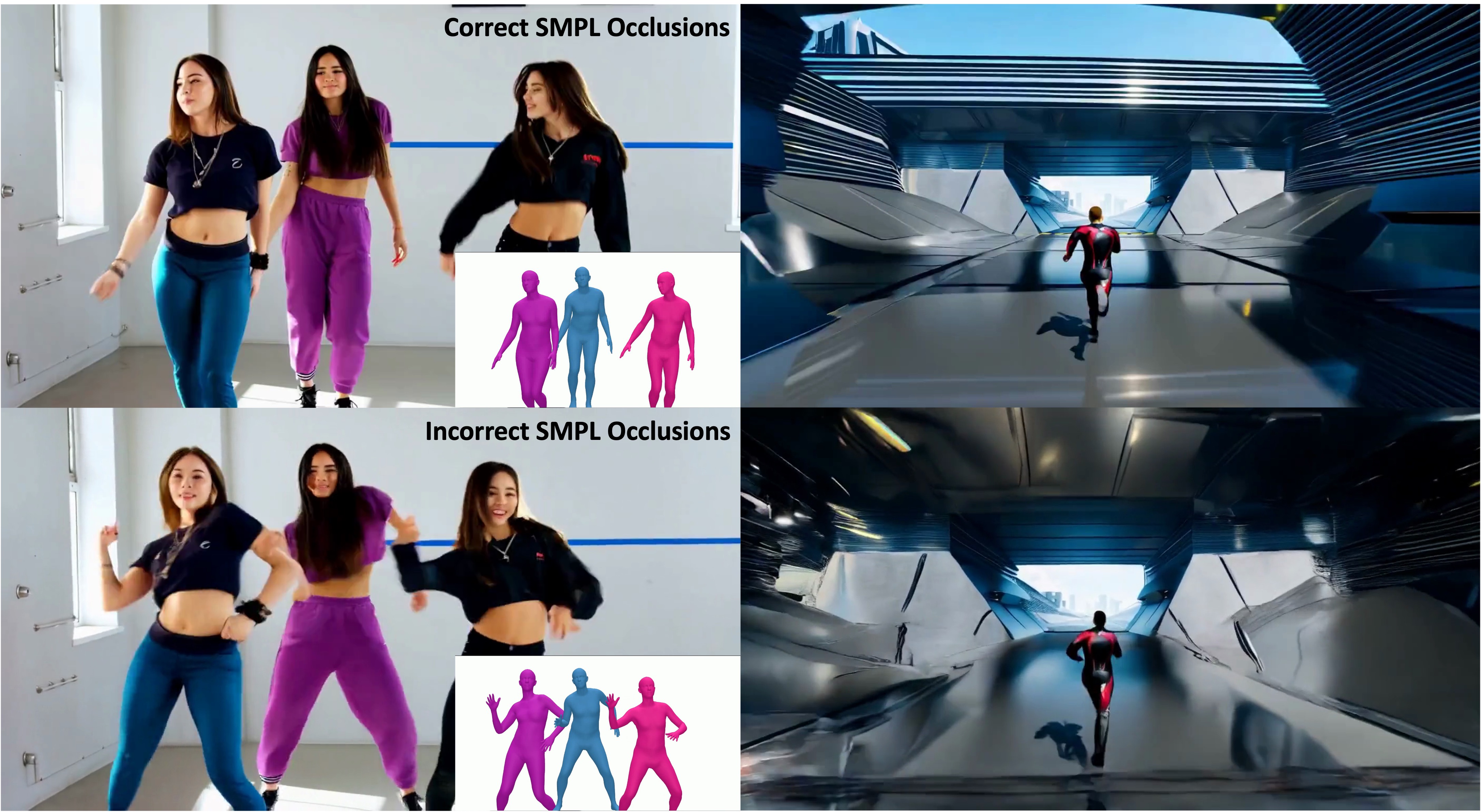}
    \caption{\revise{\textbf{Failure cases.} Our method generates photorealistic multi-character videos given accurate SMPL estimations (top left), but fails when inter-character occlusions are incorrectly estimated (bottom left). Rapid camera movements also introduce background distortions, as shown on the right (top: reference image; bottom: results with aggressive viewpoint changes).}}
    \label{fig:failure}
\end{figure}

\section{Discussion}

\paragraph{\bf{Limitations.}} \revise{While our approach successfully generates natural and realistic videos from imprecise SMPL estimates, it still relies on these SMPL estimations as input. In complex scenarios with multiple interacting people, errors in estimating occlusion relationships between SMPL models can lead to significant artifacts in the generated videos, as shown in \cref{fig:failure}}. Additionally, although our model can generate camera-controllable human videos, it struggles with cases involving extreme camera variations. This limitation is particularly evident in background generation across wide viewpoint ranges as shown in \cref{fig:failure}. The challenge of generating consistent and realistic backgrounds across 360 degrees is substantial, requiring significantly larger models and extensive video training datasets to achieve satisfactory results. 

\paragraph{\revise{\bf{Ethics Considerations.}}}

\revise{Our research presents advanced generative AI capabilities for human video synthesis. We firmly oppose the misuse of our technology for generating manipulated content of real individuals. While our model enables the creation and editing of photorealistic digital humans, we strongly condemn any application aimed at spreading misinformation, damaging reputations, or creating deceptive content. We acknowledge the ethical considerations surrounding this technology and are committed to responsible development and deployment that prioritizes transparency and prevents harmful applications.}

\paragraph{\bf{Conclusion.}} In summary, we introduce a novel and scalable interspatial attention (ISA) mechanism that seamlessly integrates with modern, scalable diffusion transformers to address the challenges of controllable photorealistic 4D human video generation. Through the combination of ISA, which leverages specialized 3D--2D relative positional encodings, and a custom video VAE, our approach achieves a significantly higher quality and consistency than baselines. Our model's ability to maintain precise control over camera and human poses while generating high-quality videos of multiple humans represents a significant advancement in the field of human video generation.

\paragraph{\bf{Acknowledgment.}} We would like to thank Shengqu Cai, He Hao, and Kecheng Zheng for their valuable discussions and insightful suggestions throughout the development of this work. We also thank all co-authors for their contributions and collaboration on this project. This work was partially supported by Google, and we gratefully acknowledge their support.

\bibliographystyle{ACM-Reference-Format}
\bibliography{reference}

\clearpage

\end{document}